\documentclass[runningheads]{llncs}
\usepackage{graphicx}

\usepackage{tikz}
\usepackage{comment}
\usepackage{amsmath,amssymb} 
\usepackage{color}

\usepackage[accsupp]{axessibility}  

\usepackage{makecell}
\usepackage{subcaption}

\usepackage{booktabs}
\usepackage[natbib=true,style=lncs]{biblatex}
\usepackage[american]{babel}
\usepackage{csquotes}

\usepackage{hyperref}
\usepackage{url}
\usepackage{multirow}

\usepackage{todonotes}

\usepackage[capitalize]{cleveref}
\crefname{section}{Sec.}{Secs.}
\crefname{section}{Section}{Sections}
\crefname{table}{Table}{Tables}
\crefname{table}{Tab.}{Tabs.}

\usepackage[export]{adjustbox}

\DeclareMathOperator*{\argmin}{argmin}
\newcommand{\VC}{\textsc{vqgan-clip}}

\begin{document}
\pagestyle{headings}
\mainmatter
\def\ECCVSubNumber{8048}  

\title{VQGAN-CLIP: Open Domain Image Generation and Editing with Natural Language Guidance} 

\titlerunning{VQGAN-CLIP: Open Domain Image Generation}
%

\author{Katherine Crowson\thanks{Co-first authors}\inst{1} \and
Stella Biderman${}^\ast$\inst{1,2} \and
Daniel Kornis \inst{3} \and
Dashiell Stander\inst{1} \and
Eric Hallahan\inst{1} \and
Louis Castricato\inst{1,4} \and
Edward Raff\inst{2}}
\authorrunning{K Crowson, S Biderman et al.}
%
\institute{EleutherAI \and Booz Allen Hamilton \and AIDock \and Georgia Institute of Technology}
\maketitle

\begin{abstract}
Generating and editing images from open domain text prompts is a challenging task that heretofore has required expensive and specially trained models. We demonstrate a novel methodology for both tasks which is capable of producing images of high visual quality from text prompts of significant semantic complexity without any training by using a multimodal encoder to guide image generations. We demonstrate on a variety of tasks how using CLIP \citep{Radford2021LearningTV} to guide VQGAN \citep{esser2021taming} produces higher visual quality outputs than prior, less flexible approaches like minDALL-E \citep{kakaobrain2021minDALL-E}, GLIDE \citep{nichol2021glide} and Open-Edit \citep{liu2020open}, despite not being trained for the tasks presented. Our code is available in a \href{https://github.com/EleutherAI/vqgan-clip/tree/main/notebooks}{public repository}.
\keywords{generative adversarial networks; grounded language; image manipulation}
\end{abstract}

\section{Introduction}

Using free-form text to generate or manipulate high-quality images is a challenging task, requiring a grounded learning between visual and textual representations. Manipulating images in an open domain context was first proposed by the seminal Open-Edit \citep{liu2020open}, which allowed text prompts to alter an image's content. This was done mostly with semantically simple transformations (e.g., turn a red apple green), and does not allow generation of images. Soon after DALL-E \citep{ramesh2021zero} and GLIDE \citep{nichol2021glide} were developed, both of which can perform generation (and inpainting) from arbitrary text prompts, but do not themselves enable image manipulation.

In this work we propose the first a unified approach to semantic image generation and editing, leveraging a pretrained joint image-text encoder \citep{Radford2021LearningTV} to steer an image generative model \citep{esser2021taming}. Our methodology works by using the multimodal encoder to define a loss function evaluating the similarity of a (text, image) pair and backpropagating to the latent space of the image generator. We iteratively update the candidate generation until it is sufficiently similar to the target text. The difference between using our technique for generation and editing is merely a matter of initializing the generator with a particular image (for editing) or with random noise (for generation).

A significant advantage of our methodology is the lack of additional training required. Only a pretrained image generator and a joint image-text encoder are necessary, while all three of \citet{liu2020open,ramesh2021zero,nichol2021glide} require training similar models from scratch. Additionally \citet{ramesh2021zero,nichol2021glide} train generators from scratch.

We demonstrate several significant contributions, including:
\begin{enumerate}
    \item High visual quality for both generation and manipulation of images. 
    \item High semantic fidelity between text and generation, especially when semantically unlikely content co-occurs. 
    \item Efficiency in that our method requires no additional training beyond the pre-trained models, using only a small amount of optimization per inference. 
    \item The value of open development and research. This technique was developed in public and open collaboration has been integral to its rapid real-world success. Non-authors have already extended our approach to other modalities (e.g., replacing text for audio) and commercial applications. 
\end{enumerate}

The rest of our manuscript is organized as follows. In \cref{sec:method} we discuss how of how our methodology works, resulting in a simple and easy-to-apply approach for combing multiple modalities for generation or manipulation. The efficacy of \VC{} in generating high quality and semantically relevant images is shown in \cref{sec:generation}, followed by superior manipulation ability in \cref{sec:editing}. The design choices of \VC{} to obtain both high image quality and fast generation are validated by ablations in \cref{sec:ablations}, and \cref{sec:resource} discusses resource usage and efficiency considerations. As our approach has been public since April 2021, we are able to show further validation by external groups in \cref{sec:useage}. This use includes extensions to other modalities, showing the flexibility of our approach, as well as commercial use of \VC{} that demonstrate its success at handling open-domain prompts and images to a satisfying degree. Finally we conclude in \cref{sec:conclusion}.

\section{Our Methodology} \label{sec:method}

To demonstrate our method's effectiveness we apply it using VQGAN \citep{esser2021taming} and CLIP \citep{Radford2021LearningTV} as pre-trained models, and so refer to our approach as \VC{}. We stress, however, that our approach is not specific to either model and that subsequent work has already shown success that builds on our work using other models \citep{chendiffvg+,fei2021wenlan,michel2021text2mesh,tian2021modern}, and even in other modalities \citep{jang2022music2video,wu2021wav2clip}.

\begin{figure*}[!hbt]
  \centering
  \includegraphics[width=\textwidth]{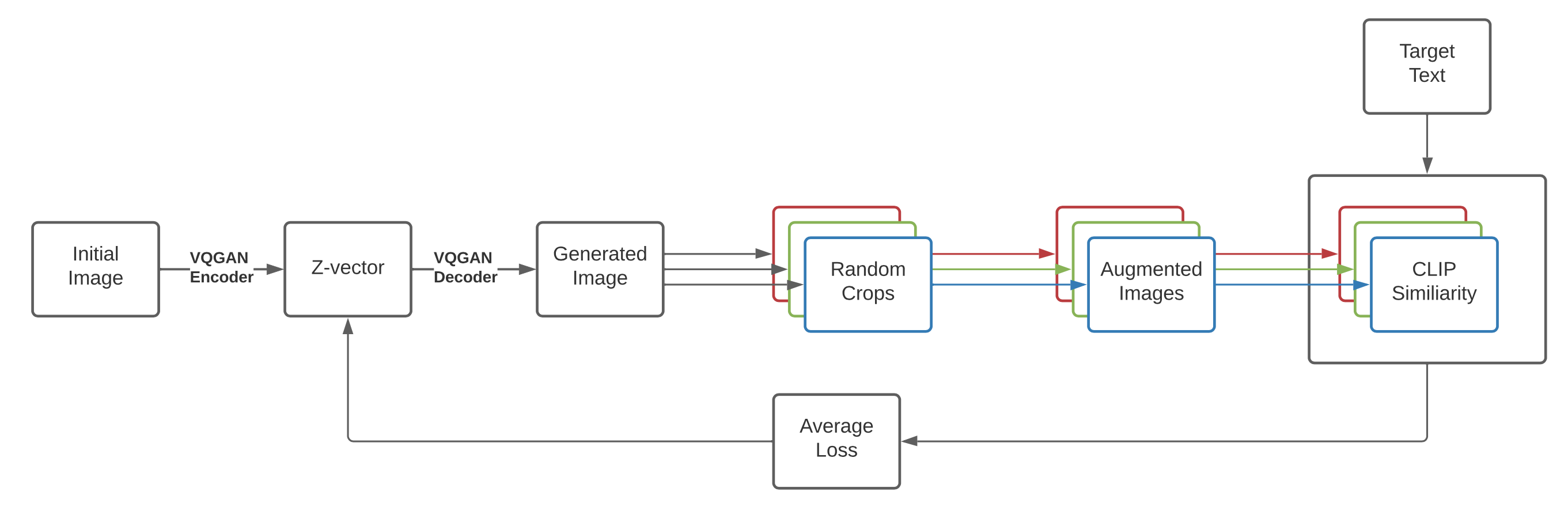}
  \caption{Diagram showing how augmentations are added to stabilize and improve the optimization. Multiple crops, each with different random augmentations, are applied to produce an average loss over a single source generation. This improves the results with respect to a single latent Z-vector.}\label{fig:method}
\end{figure*}

We start with a text prompt and use a GAN to iteratively generate candidate images, at each step using CLIP to improve the image. We optimize the image by treating the squared spherical distance between the embedding of the candidate and the embedding of the text prompt as a loss function, and differentiating through CLIP with respect to the GAN's latent vector representation of the image, which we refer to as the ``z-vector'' following \citet{van2017neural}. This process is outlined in \cref{fig:method}

To generate an image, the ``initial image'' contains random pixel values. The optimization process is repeated to alter the image, until the output image gradually improves such that it semantically matches the target text. We can also edit existing images by starting with the image-to-edit as the ``initial image''. The text prompt used to describe how we want the image to change is used identically to the text prompt for generating an image, and no changes to the architecture exist between generation and manipulation besides how the `initial image'' is selected.

We use Adam \citep{kingma2014adam} to do the actual optimization, a learning rate of $0.15$, $\beta = (0.9, 0.999)$, and run for 400 iterations for the experiments in this paper.

\subsection{Discrete Latent Spaces for Images}

Unlike the naturally discrete nature of text, the space of naturally occurring images is inherently continuous and not trivially discretized. Prior work by \citet{van2017neural} borrows techniques from vector quantization (VQ) to  represent a variety of modalities with discrete latent representations by building a codebook vocabulary with a finite set of learned embeddings. Given a codebook of vocabulary size $K$ with embedding dimension $n_k$, $\mathcal{Z} = \{z_i\}_k^K \in \mathbb{R}^{n_k}$.

This is applied to images by constructing a convolutional autoencoder with encoder $\mathcal{E}$ and decoder $\mathcal{G}$. An input image $x \in I$ is first embedded with the encoder $z = E(x)$. We can then compute the vector quantized embedding $x$ as $$z_q = \argmin_{z_k \in \mathcal{Z}} \|z_{i,j} - z_k\|$$
which we can then multiply back through the vocabulary in order to perform reconstruction. We can then use a straight-through estimator on the quantization step in order to allow the CNN and codebook to be jointly trained end-to-end. We use the popular VQGAN \citep{esser2021taming} model for the experiments in this paper.

\subsection{Contrastive Text-Image Models}

To guide the generative model, we need a way to adjust the similarity of a candidate generation with the guidance text. To achieve this, we use CLIP, \citep{Radford2021LearningTV}, a joint text-image encoder trained by using contrastive learning. We use CLIP to embed text prompts and candidate generated images independently and measure the cosine similarity between the embeddings. This similarity is then reframed as a loss that we can use gradient descent to minimize.

\subsection{Augmentations}

One challenge of using \VC{} is that gradient updates from the CLIP loss are quite noisy if calculated on a single image. To overcome this we take the generated candidate image and modify it many times, producing a large number of augmented images. We take random crops of the candidate image and then apply further augmentations such as flipping, color jitter, noising, etc. \citep{eriba2019kornia}. Most high level semantic features of an image are relatively invariant to these changes, so averaging the CLIP loss with respect to all of the augmented images reduces the variance of each update step. There is a risk that a random crop might dramatically change the semantic content of an image (e.g. by cropping out an important object), but we find that in practice this does not cause any issues.

For the results presented in this paper we used an augmentation pipeline consisting of: random horizontal flips, random affine projections, random perspective projections, random color jitter, and adding random Gaussian noise. 

\subsection{Regularizing the Latent Vector}\label{sec:regularize}

When using an unconstrained VQGAN for image generation, we found that outputs tended to be unstructured. Adding augmentations helps with general coherence, but the final output will often still contain patches of unwanted textures. To solve this problem we apply a weighted $L^2$ regularization to the the z-vector.

This produces a regularized loss function given by the equation $$\mathit{Loss}=\mathit{L}_\mathit{CLIP}+\alpha\cdot \frac{1}{N}\sum_{i=0}^{N}Z_{i}^{2}$$ where $\alpha$ is the regularization weight. This encourages parsimony in the representation, sending low information codes in VQGAN's codebook to zero. In practice we note that regularization appears to improve the coherence of the output and produces a better structured image. We decay the regularization term by $0.005$ over the course of generation.

\subsection{Additional Components}\label{sec:additional}

Our methodology is highly flexible and can be extended straightforwardly depending on the use-case and context due to the ease of integrating additional interventions on the intermediate steps of image generation. Researchers using our framework have introduced a number of additional components, ranging from using ensembles \citep{couairon2022flexit}, to using B\'ezier curves for latent representations \citep{frans2021clipdraw,chendiffvg+}, to using perturbations to make the results more robust to adversaries \citep{liu2021fusedream}. Although they aren't used in the main experiments of this paper, we wish to call attention to two in particular that we use frequently: ``prompt addition'' and masked image editing. We give an overview of both here, and provide additional experiments and information in \cref{app:more-components}

\paragraph{Prompt Addition:} We have found that our users are often interested in applying multiple text prompts at the same time. This can be achieved by computing the loss against multiple target texts simultaneously and adding the results. In \cref{app:prompt-add} we use this tool to explore the semantic cohesion of \VC{}'s generations.

\paragraph{Masking:} A common technique in image generation and editing is \textit{masking}, where a portion of an image is identified ahead of time as being where a model should edit\footnote{In the context of image this is often referred to as ``infilling,'' but we will use ``masking'' as a general term to refer to both.} \VC{} is compatible with masking by zeroing out the gradients in parts of the latent vector that one wishes to not change. However \VC{} can also leverage the semantic knowledge of CLIP to perform \textit{self-masking} without any non-textual human input. 

\section{Semantic Image Generation} \label{sec:generation}

\begin{figure}
    \centering
    \subfloat[Oil painting of a candy dish of glass candies, mints, and other assorted sweets]{\includegraphics[width=0.25\textwidth]{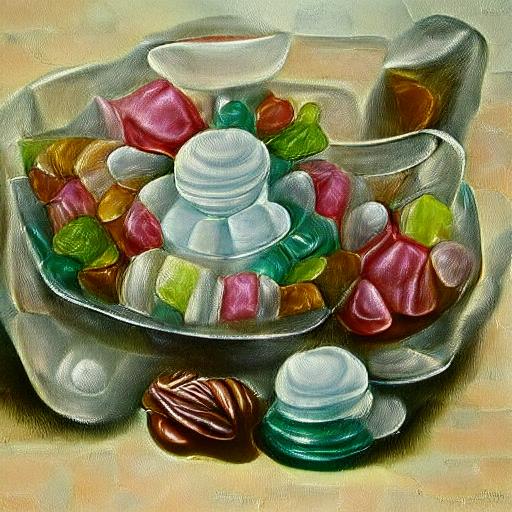}}\hfill
    \subfloat[A colored pencil drawing of a waterfall]{\includegraphics[width=0.25\textwidth]{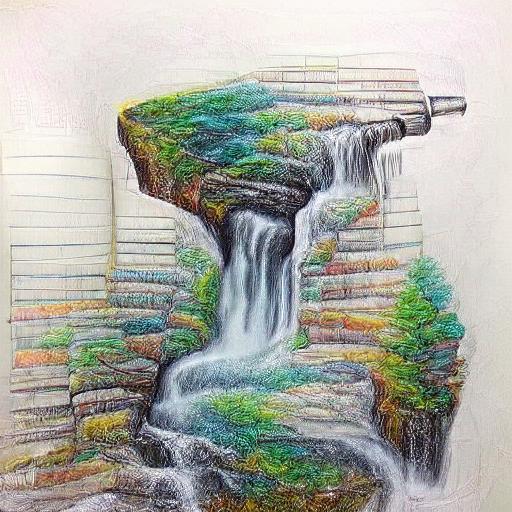}}\hfill
    \subfloat[A fantasy painting of a city in a deep valley by Ivan Aivazovsky]{\includegraphics[width=0.25\textwidth]{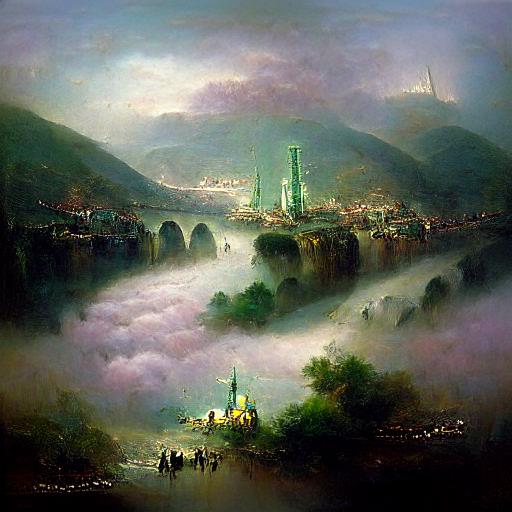}}\\
    \subfloat[A beautiful painting of a building in a serene landscape ]{\includegraphics[width=0.25\textwidth]{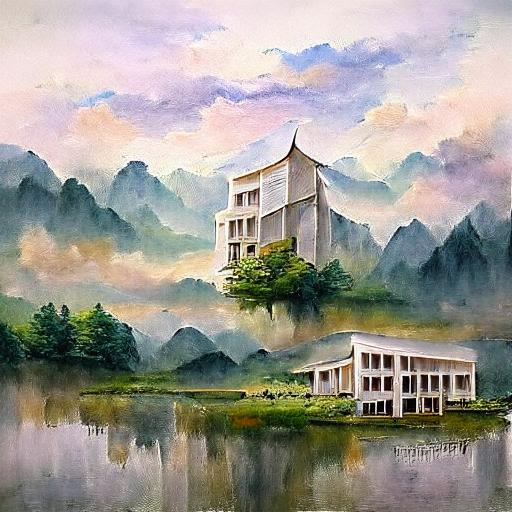}}\hfill
    \subfloat[sketch of a 3D printer by Leonardo da Vinci]{\includegraphics[width=0.25\textwidth]{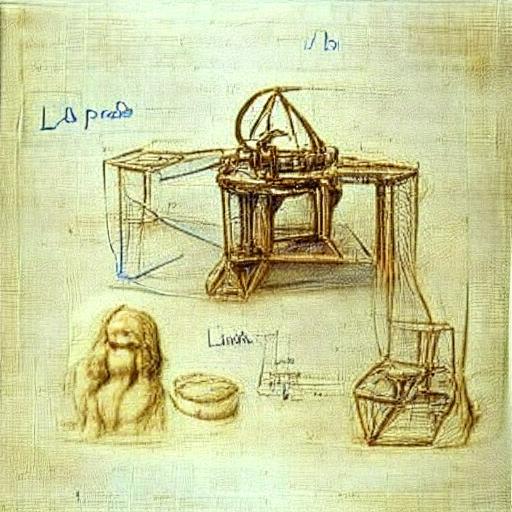}}\hfill
    \subfloat[an autogyro flying car, trending on artstation]{\includegraphics[width=0.25\textwidth]{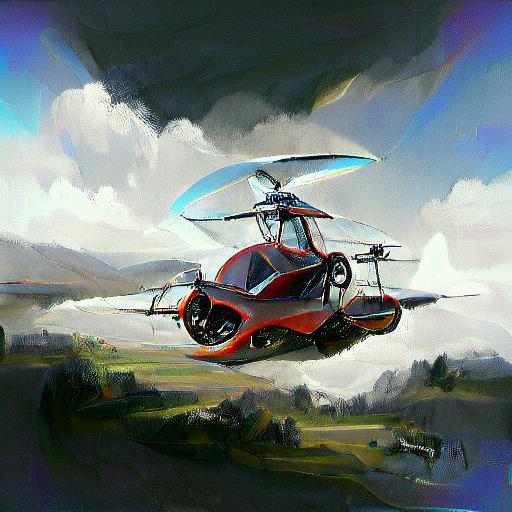}}\\
    \subfloat[an astronaut in the style of van Gogh]{\includegraphics[width=0.25\textwidth]{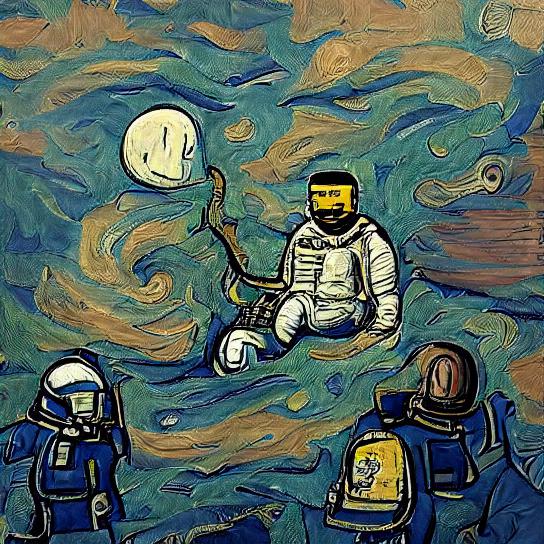}}\hfill
    \subfloat[Baba Yaga's house + fantasy art]{\includegraphics[width=0.25\textwidth]{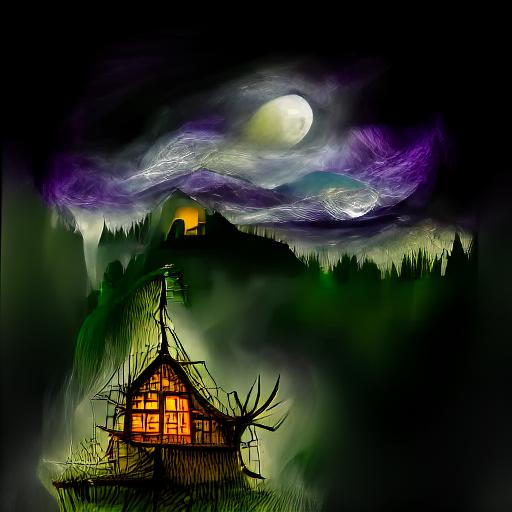}}\hfill
    \subfloat[pickled eggs, tempera on wood]{\includegraphics[width=0.25\textwidth]{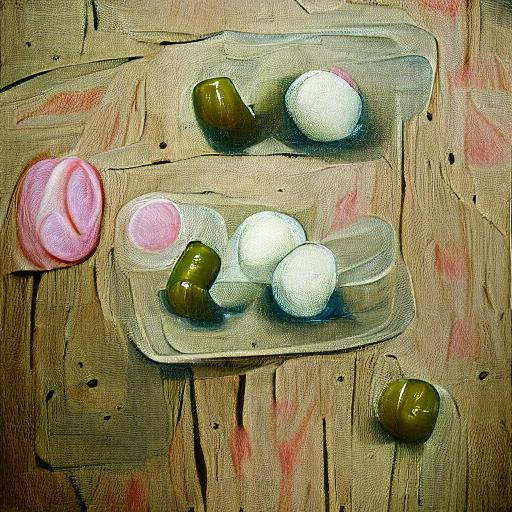}}\\\
    \subfloat[effervescent hope]{\includegraphics[width=0.25\textwidth]{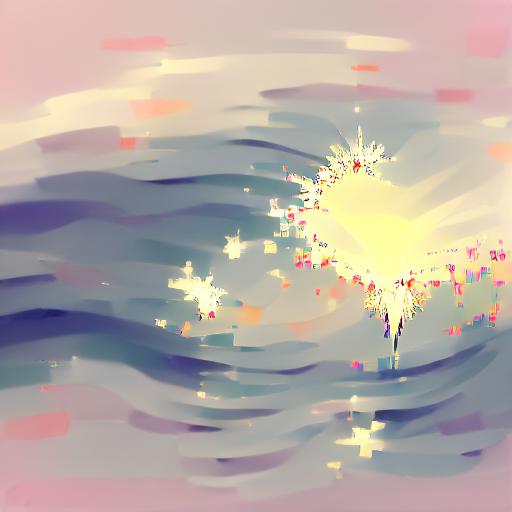}}\hfill
    \subfloat[the Tower of Babel by J.M.W. Turner]{\includegraphics[width=0.25\textwidth]{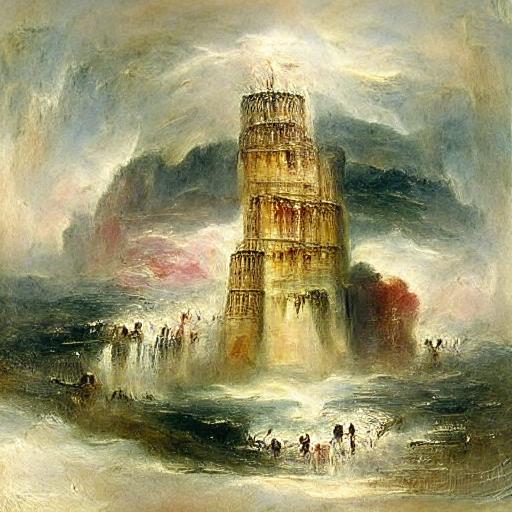}}\hfill
    \subfloat[a futuristic city in synthwave style]{\includegraphics[width=0.25\textwidth]{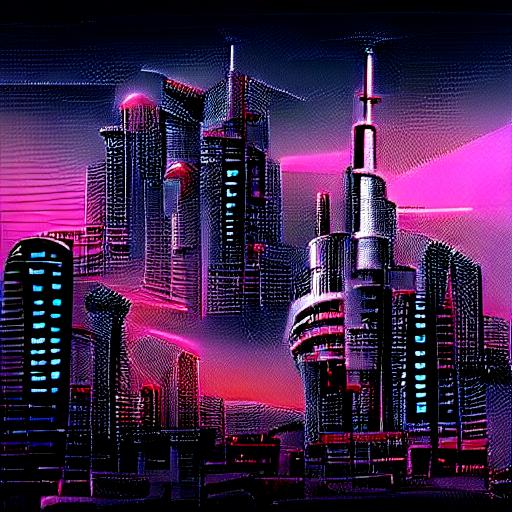}}
    \caption{Example \VC{} generations and their text prompts. Prompts selected to demonstrate a range of visual styles that \VC{} is capable of producing including classical art (g, i), modern art (l), drawings (e), oils (a), and others not included due to space.}
    \label{fig:example-gen}
\end{figure}

The primary application of our methodology is for generating images from text. In contrast to previous work on this topic \citep{ramesh2021zero,nichol2021glide,wang2022clip}, we do not perceive creating photo-realistic images or images that could convince a human that they are real photographs as our primary goal. Our focus is on producing images of high visual quality that are semantically meaningful in relation to a natural language prompt, which we demonstrate in this section. This in fact requires abandoning photo-realism when prompts may ask for artistic or explicitly unrealistic generations and edits. 
A loosely curated set of example generations is presented in \cref{fig:example-gen}.

As \VC{} has been publicly available for almost a year, we have had the opportunity to observe people experimenting with and building off of \VC{} in the wild. In \cref{app:generations} we show a sample of artwork created by people other than the authors of this paper  are included to demonstrate the power and range of \VC{}\!.

\subsection{Artistic Impressions}\label{sec:famous}

\begin{figure*}
    \centering
    \subfloat[van Gogh]{\includegraphics[width=0.25\textwidth]{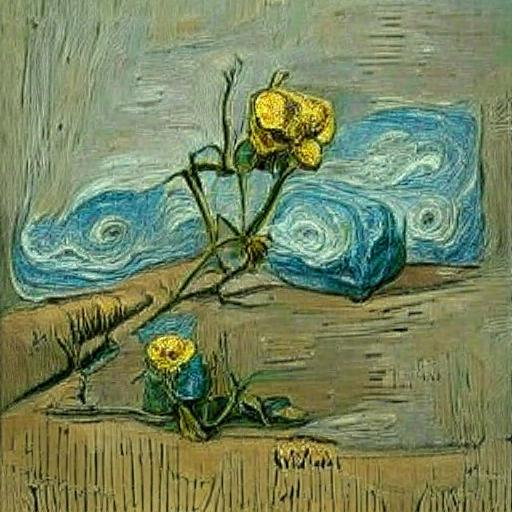}}\hfill
    \subfloat[Picasso]{\includegraphics[width=0.25\textwidth]{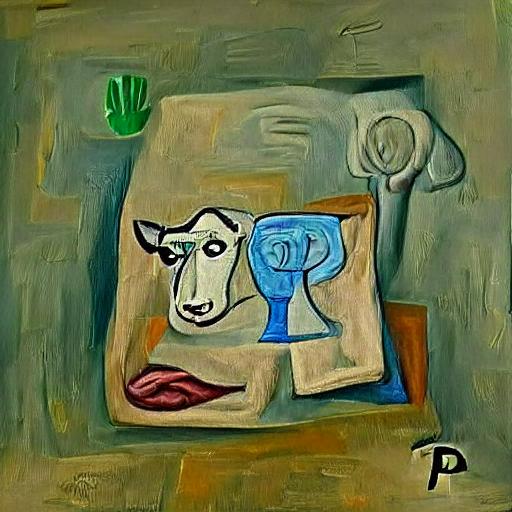}}\hfill
    \subfloat[Hokusai]{\includegraphics[width=0.25\textwidth]{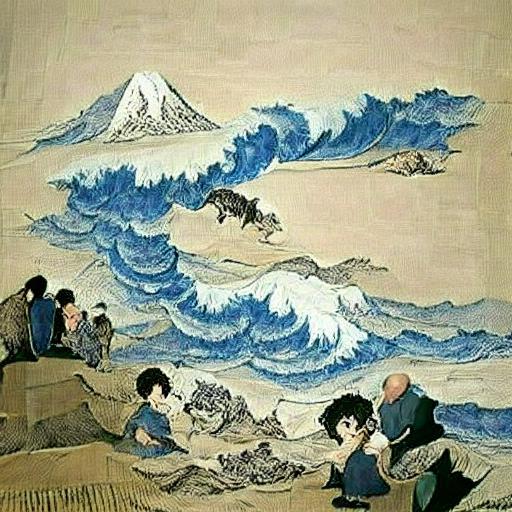}}\\
    \subfloat[Turner]{\includegraphics[width=0.25\textwidth]{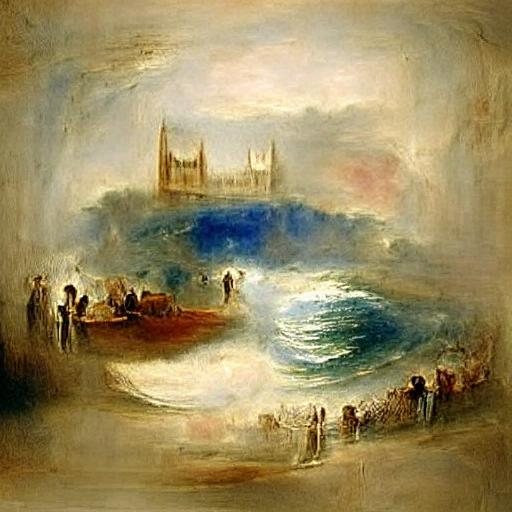}}\hfill
    \subfloat[Kahlo]{\includegraphics[width=0.25\textwidth]{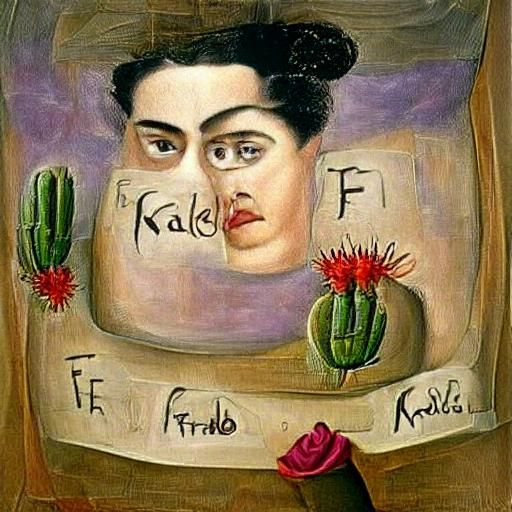}}\hfill
    \subfloat[Mehretu]{\includegraphics[width=0.25\textwidth]{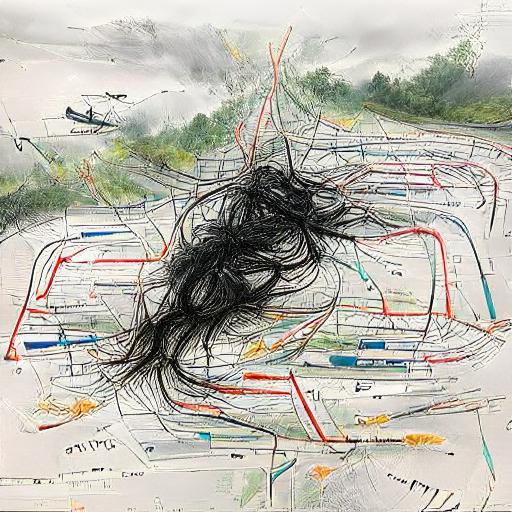}}\\
    \caption{Stylistic impressions of famous artists. Third party tools like Google's Reverse Image Search indicate that real paintings by the target artists are the most visually similar images in every case.}
    \label{fig:famous-art}
\end{figure*}

We find that \VC{} is able to evoke the artistic style of famous artists and major artistic styles from around the world. \cref{fig:example-gen} features ``an astronaut in the style of van Gogh'' whose background evokes Starry Night and ``the Tower of Babel by J. M. W. Turner'' which draws on Turner's color palate and use of light. Another way this can be seen is by directly asking for ``a painting by [name]'' or ``art by [name].'' In \cref{fig:famous-art} we present six images created this way drawing on artists from different regions, time periods, and artistic styles. While the images often are missing cohesion (most likely due to the vagueness of ``a painting'' as a prompt) they are each markedly reminiscent of the artist in question. While CLIP would obviously find these images visually similar to other works by the artist, we also find that non-CLIP-based image similarity approaches reliably identify these images as visually similar to work by the artists. To validate this we queried Google's Reverse Image Search using each generation in \cref{fig:famous-art}, and in every case a real painting by the target artist was the most similar image.

\subsection{Comparisons to Other Approaches}

The closest prior work in open domain generation of images comes from DALL-E \citep{ramesh2021zero} and GLIDE \citep{nichol2021glide}, which claim to train very large pretrained text-to-image models. DALL-E and GLIDE are purported to be 12 billion and 5 billion parameters, respectively, while \VC{} together is 227 million. Unfortunately, we were denied permission to study the models purported in the respective papers by their authors. We instead use the verifiable state-of-the-art models using each methodology methodologies. This includes the independent implementation minDALL-E \cite{kakaobrain2021minDALL-E} (1.3 B parameters) and two versions of GLIDE (783 M parameters without CLIP and 941 M with) released by OpenAI\footnote{It is important to stress that at time of writing there was no verifiable evidence for the existence or performance of either of these models other than the word of OpenAI employees. That such assertions are taken seriously as science is a symptom of the pervasive power disparities across machine learning research \citep{black2022gpt,leahy2021hard,miceli2022studying,tao2021insiders}}.

To evaluate our model, we recruited humans and asked them to rate the alignment of (text, image) pairs on a scale of 1 (low) to 5 (high). In particular, they were directed to rate higher quality images that do not match the prompt lower than lower quality images that do. Prompts were selected based on principles learned from our experience working with these models but without prior knowledge of how the models would behave on the particular prompts in question. All prompts and generated images can be found in \Cref{app:generations}. To provide the maximal advantage to our competitors, minDALL-E and GLIDE examples are cherry-picked best-of-five, while \VC{} examples are uncherry-picked (best-of-one). Table 1 shows the mean score per prompt for each model. We find that humans overwhelmingly view the generations using our technique as more aligned with the input text.

\begin{table}[h]
    \centering
    \begin{tabular}{l|c c c c c c c c c c c c c}
                                & A   & B    & C   & D & E & F & G & H & I & J & K & L & \textbf{Mean}\\\midrule
        minDALL-E     & 3.3 & 2.3 & 3.2 & 3.7 & 1.5 & 2.7 & 2.2 & 1.3 & 3.3 & 3.0 & 3.5 & 2.3 & 2.7\\\
        GLIDE (CF)    & 3.0 & 4.0 & 2.7  & 3.2 & 1.3 & 2.5 & 1.3 & 1.2 & 2.0 & 2.3 & 2.0 & 2.8 & 2.3\\\
        GLIDE (CLIP) & 3.2 & 4.0 & 2.8  & 4.8 & 3.0 & 2.7 & 1.8 & 2.5 & 3.7 & 2.3 & 3.7 & \textbf{5.0} & 3.3\\\
        VQGAN-CLIP & \textbf{4.3} & \textbf{5.0} & \textbf{4.8} & \textbf{5.0} & \textbf{4.5} & \textbf{4.5} & \textbf{4.8} & \textbf{4.7} & \textbf{4.2} & \textbf{3.8} & \textbf{4.7} & 4.7 & \textbf{4.6}\\\midrule
    \end{tabular}
    \caption{Mean human ratings of generations by each model on a score of 1 (worst) to 5 (best).}
    \label{tab:human}
\end{table}\vspace{-0.2in}

\subsection{Qualitative Analysis}

\begin{figure}[!htb]
    \centering
    \subfloat{\includegraphics[width=0.2\textwidth]{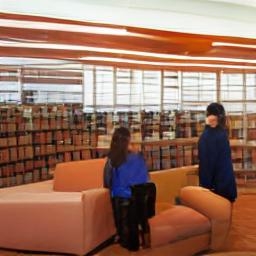}}\vspace{-00pt}\hfill
    \subfloat{\includegraphics[width=0.2\textwidth]{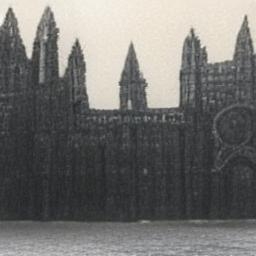}}\hfill
    \subfloat{\includegraphics[width=0.2\textwidth]{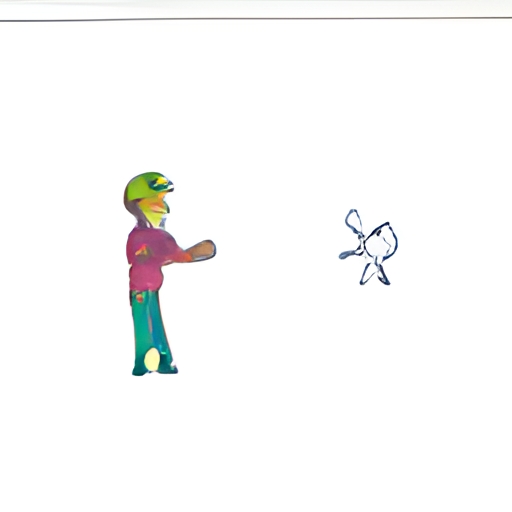}}\hfill
    \subfloat{\includegraphics[width=0.2\textwidth]{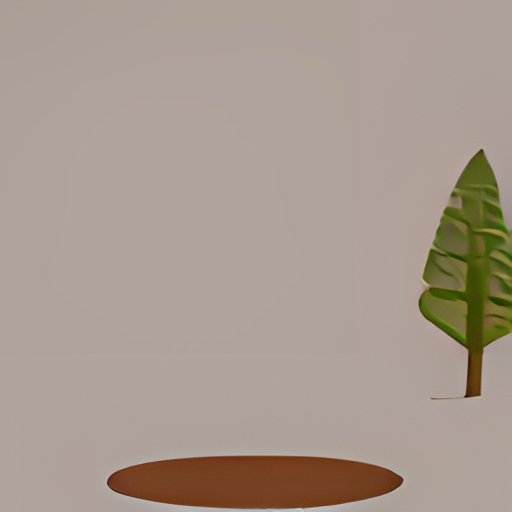}}\\
    \subfloat{\includegraphics[width=0.2\textwidth]{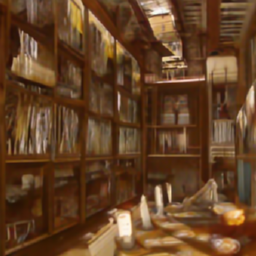}}\vspace{-00pt}\hfill
    \subfloat{\includegraphics[width=0.2\textwidth]{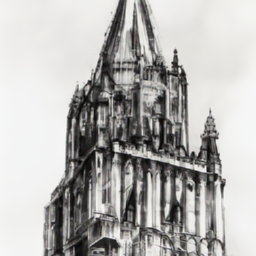}}\hfill
    \subfloat{\includegraphics[width=0.2\textwidth]{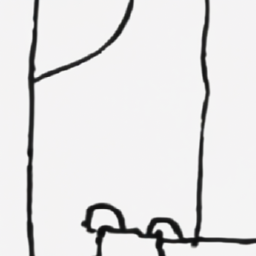}}\hfill
    \subfloat{\includegraphics[width=0.2\textwidth]{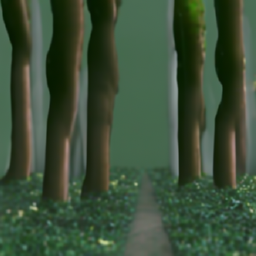}}\\
    \subfloat{\includegraphics[width=0.2\textwidth]{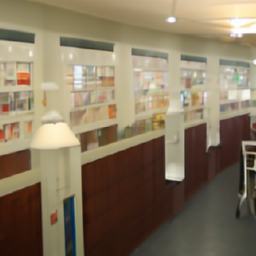}}\vspace{-00pt}\hfill
    \subfloat{\includegraphics[width=0.2\textwidth]{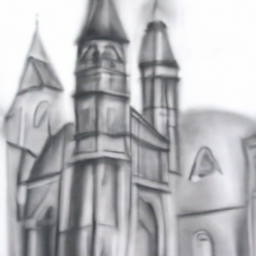}}\hfill
    \subfloat{\includegraphics[width=0.2\textwidth]{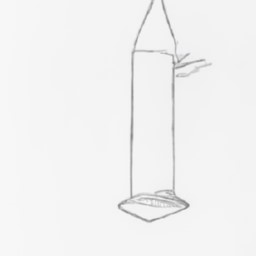}}\hfill
    \subfloat{\includegraphics[width=0.2\textwidth]{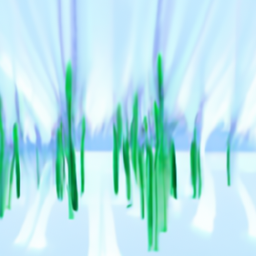}}\\
    \setcounter{subfigure}{0}
    \subfloat[the universal library trending on artstation]{\includegraphics[width=0.2\textwidth]{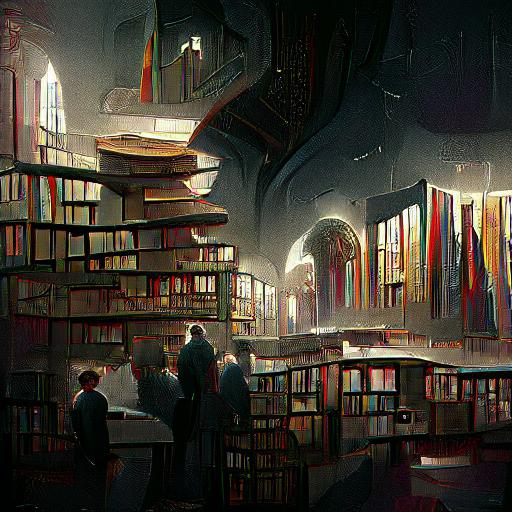}}\hfill
    \subfloat[a charcoal drawing of a cathedral]{\includegraphics[width=0.2\textwidth]{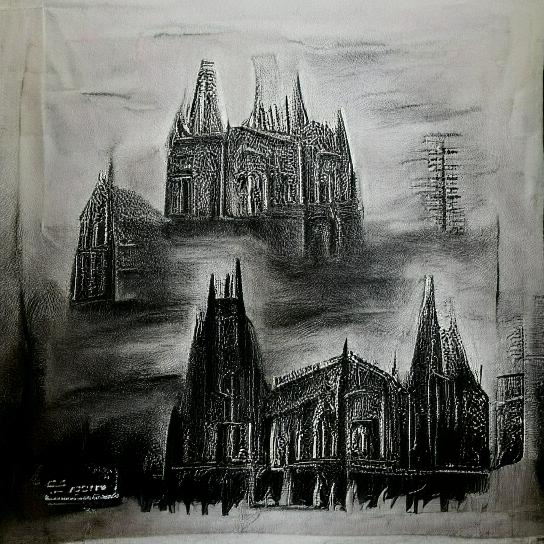}}\hfill
    \subfloat[a child’s drawing of a baseball game]{\includegraphics[width=0.2\textwidth]{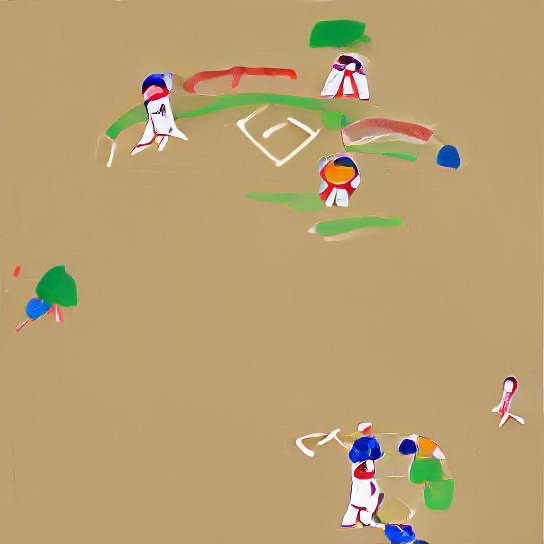}}\hfill
    \subfloat[a forest rendered in low poly]{\includegraphics[width=0.2\textwidth]{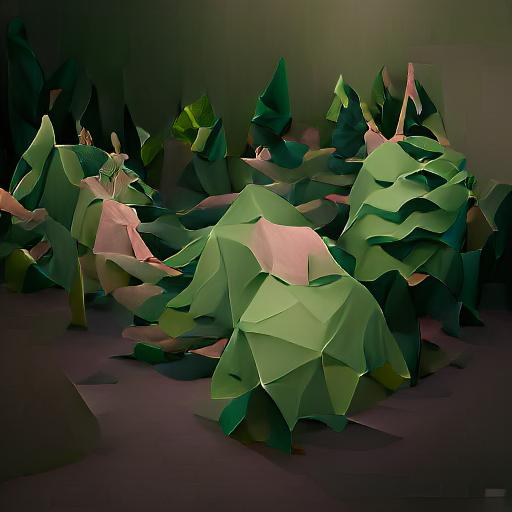}}
    \caption{Text based generations of images. Top to bottom: minDALL-E, GLIDE (CLIP-guided), GLIDE (CF-guided), and our \VC{}.}
    \label{fig:generation-comparison-basic}
\end{figure}

A sampling of representative results is shown in \cref{fig:generation-comparison-basic} for four different prompts using minDALL-E, two variants of GLIDE, and our \VC{}.  Further comparisons, including the prompts in \cref{fig:example-gen}, can be found in \cref{app:comparisons}. We find that the minDALL-E and the GLIDE models are much more variable in the quality of their generations. While they are able to produce images that are clearly recognizable in response to the prompts ``the universal library trending on artstation'' and ``a charcoal drawing of a cathedral,'' their generations in response to ``a child’s drawing of a baseball game'' are largely unrecognizable and their responses to ``a forest rendered in low poly'' ignore the later half of the prompt. These latter cases demonstrate the low semantic relevance of prior methods' output given the prompt.

The ``child's drawing'' case is of particular note here in that a child's drawing is expected to have lower visual clarity and lack of structure. That \VC{} is able to correctly modulate its ability for fine details is thus of note to show that \VC{} is not intrinsically biased toward producing fine details when inappropriate, and correctly identifies the appropriate context of multi-part prompts. Further evidence of this can be found in \cref{fig:example-gen-min,fig:example-gen-clip,fig:example-gen-glide,fig:example-gen-vc} where \VC{} is able to produce generations for the prompts ``A colored pencil drawing of a waterfall'' and ``sketch of a 3D printer by Leonardo da Vinci'' that showcase the properties of the medium (visible strokes, the use of shading, the fact that the image is created on a piece of paper) while still producing a compelling visual image.

\section{Semantic Image Editing} \label{sec:editing}

As far as we are aware, our framework is the first in the literature to be able to perform semantic image generation \textit{and} semantic image editing. There are other examples in the literature of generative models that can perform style transfer \citep{patashnik2021styleclip}, image inpainting \citep{nichol2021glide,ramesh2021zero}, and other types of image manipulation \citep{ntavelis2020sesame}, but we note that each of these represent distinct tasks from open domain semantic image editing. By contrast, to adapt our generation methodology to image editing all that is required is to replace the randomly initialized starting image with the image we wish to edit.

\subsection{Comparison to State-of-the-Art}

For semantic image editing we compare to Open-Edit \citep{liu2020open}. As far as we are aware, Open-Edit is the only published research on \textit{open domain} semantic image editing other than our work. To avoid giving any accidental advantage to our methodology, we focus primarily on the domains presented as examples in \citet{liu2020open} such as changing colors and textures. We use the default settings for their model and the same prompting structure as in their paper.

\subsubsection{Color editing} 
\begin{figure}[!h]
\centering
\adjustbox{max width=\columnwidth}{%
\begin{tabular}{cccc}
Instruction & Original &  VQGAN-CLIP & Open-Edit\\
  ``Green''
& \includegraphics[width=.2\linewidth,valign=m]{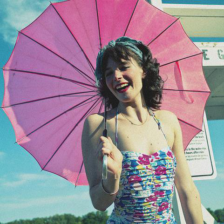}
& \includegraphics[width=.2\linewidth,valign=m]{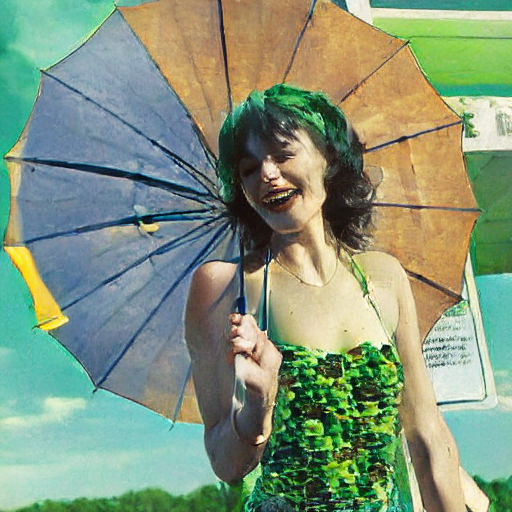}
& \includegraphics[width=.2\linewidth,valign=m]{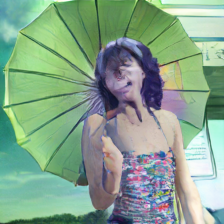}\\
  ``Red Bus''
& \includegraphics[width=.2\linewidth,valign=m]{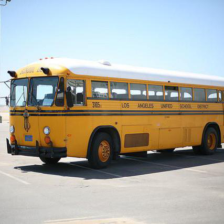}
& \includegraphics[width=.2\linewidth,valign=m]{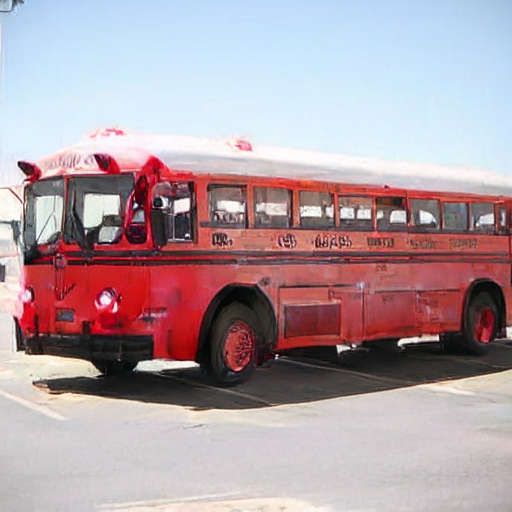}
& \includegraphics[width=.2\linewidth,valign=m]{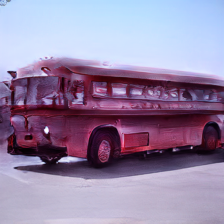}
\end{tabular}
}
\caption{Examples of editing the color in an image. Original on the left, our \VC{} in the middle, and Open-Edit on the right. \VC{} better maintains original structure of the content while limiting unintended distortion.} \label{tbl:color}
\end{figure}

Here we prompt the model to change the dominant color palette without degrading the image quality or any of the finer details. The results can be seen in \cref{tbl:color}, where prior Open-Edit causes destructive transformations of the content of the image. In the second case the ``Red bus'' also shows a single desired target for manipulation that is respected by \VC, but Open-Edit causes a change in coloration of the entire image.

\subsubsection{Weather Modification} 
\begin{figure}[!ht]
    \centering
\adjustbox{max width=\columnwidth}{%
\begin{tabular}{cccc}
Instruction & Original &  VQGAN-CLIP & Open-Edit\\
``Foggy Sky $\to$ Clear Sky''
& \includegraphics[width=.2\linewidth,valign=m]{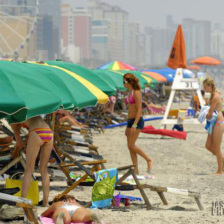}
& \includegraphics[width=.2\linewidth,valign=m]{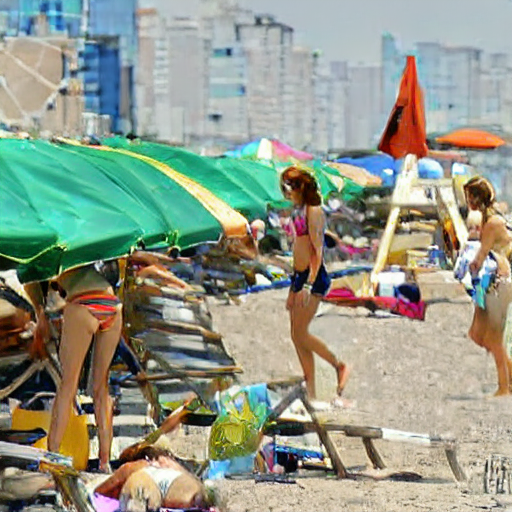}
& \includegraphics[width=.2\linewidth,valign=m]{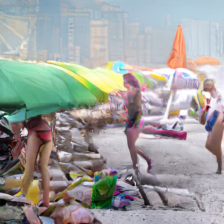}\\
``Clear Sky $\to$ Cloudy Sky''
& \includegraphics[width=.2\linewidth,valign=m]{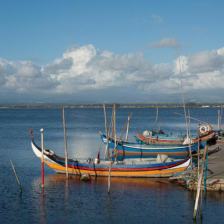}
& \includegraphics[width=.2\linewidth,valign=m]{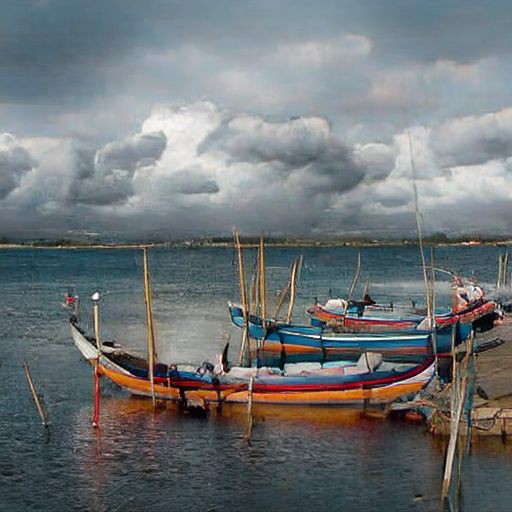}
& \includegraphics[width=.2\linewidth,valign=m]{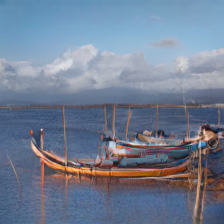}\\
``Cloudy $\to$ Sunny''
& \includegraphics[width=.2\linewidth,valign=m]{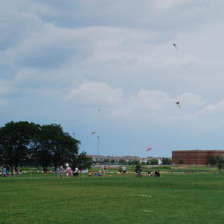}
& \includegraphics[width=.2\linewidth,valign=m]{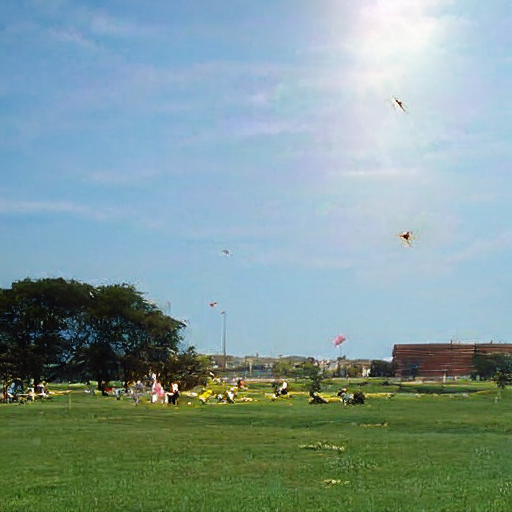}
& \includegraphics[width=.2\linewidth,valign=m]{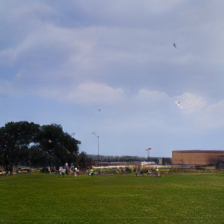}
\end{tabular}
}
\caption{Weather alteration can required greater alteration of scene structure that Open-Edit is not able to perform, as shown in the ``Cloudy $\to$ Sunny'' example that needs to alter the sky in addition to brightness levels.}\label{fig:weather_mod}
\end{figure}

Another use case that \citet{liu2020open} highlight as a success of their model is weather modification, changing the overall weather conditions present in an image. Results on this task are shown in \cref{fig:weather_mod}, where Open-Edit's reliance on edge maps to maintain structure show a limitation in editing ability. The needed alterations often change more of the image content that would violate the edge maps, preventing Open-Edit from being as successful in achieving the desired content change. 

\subsubsection{Misc} 
\begin{figure}[!ht]
\centering
\adjustbox{max width=\columnwidth}{%
\begin{tabular}{cccc}
Instruction & Original &  VQGAN-CLIP & Open-Edit\\
  ``Wooden''
& \includegraphics[width=.2\linewidth,valign=m]{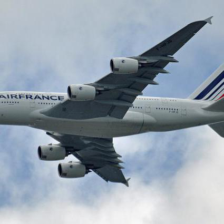}
& \includegraphics[width=.2\linewidth,valign=m]{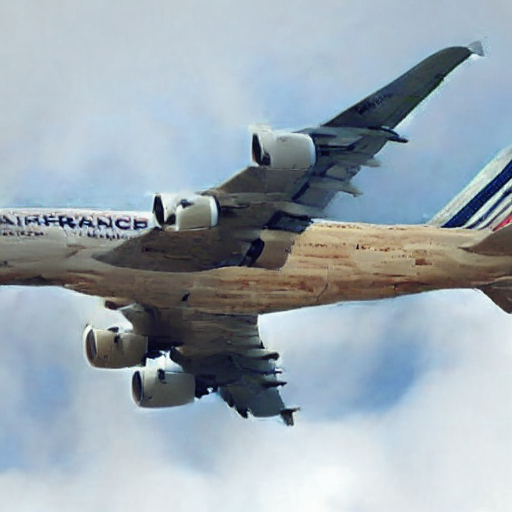}
& \includegraphics[width=.2\linewidth,valign=m]{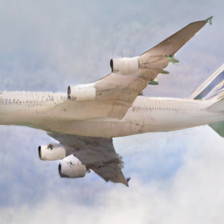}\\
  ``Withered Flowers''
& \includegraphics[width=.2\linewidth,valign=m]{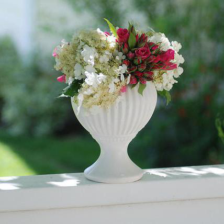}
& \includegraphics[width=.2\linewidth,valign=m]{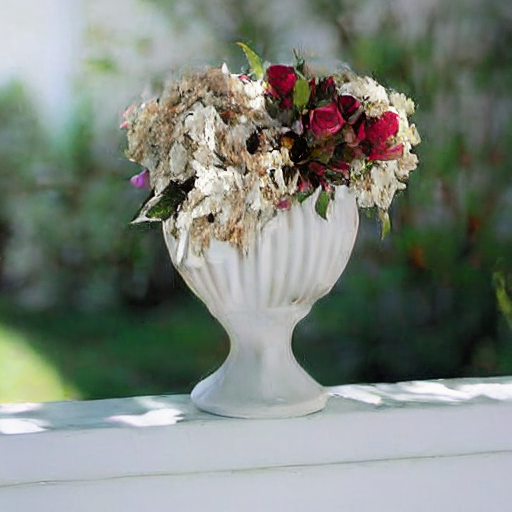}
& \includegraphics[width=.2\linewidth,valign=m]{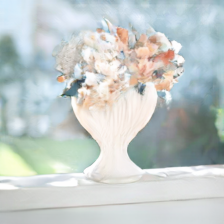}\\
  ``Focused''
& \includegraphics[width=.2\linewidth,valign=m]{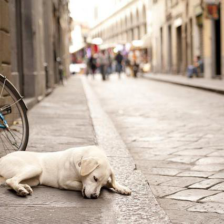}
& \includegraphics[width=.2\linewidth,valign=m]{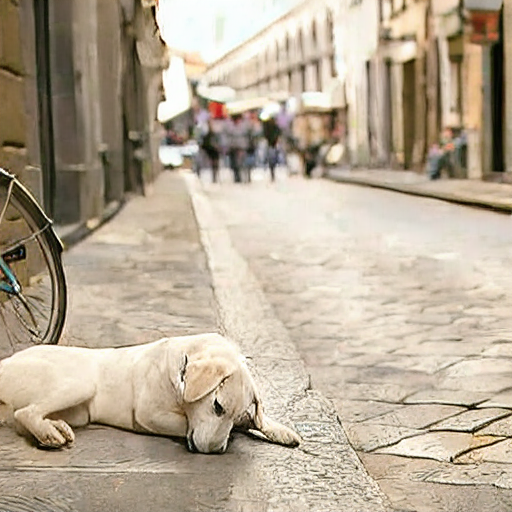}
& \includegraphics[width=.2\linewidth,valign=m]{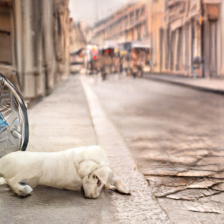}\\
\end{tabular}
}
\caption{More challenging modifications that required greater linguistic grounding to visual content to achieve, again showing \VC{} is better able to edit image content.} \label{fig:misc_edits}
\end{figure}

We include extra miscellaneous examples to emphasize that this is open domain image editing and the performance is not limited to select types of transformations. These are shown in \cref{fig:misc_edits}, and we note the ``wooden'' and ``focused'' examples demonstrate a task with less correlative semantics. This further requires a more robust grounding between modalities for success and the ability of our approach to better handle a breadth of possible inputs for open-domain prompts and images. 

\section{Ablations on Components} \label{sec:ablations}

\subsection{Latent Vector Regularization}

\begin{figure}[!h]
    \centering
    \subfloat{\includegraphics[width=0.2\textwidth]{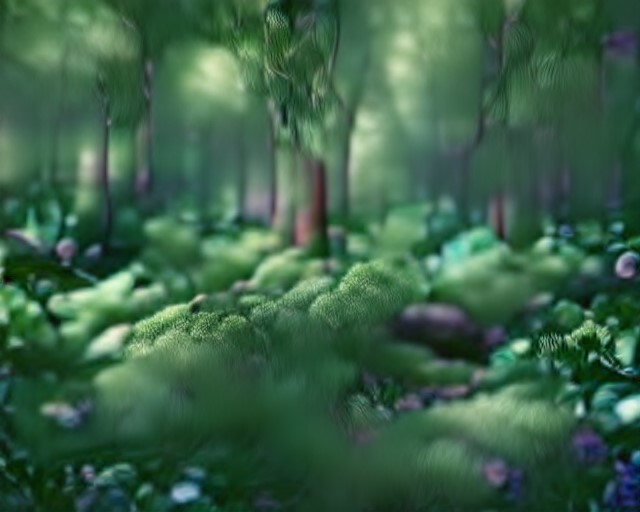}}\hfill
    \subfloat{\includegraphics[width=0.2\textwidth]{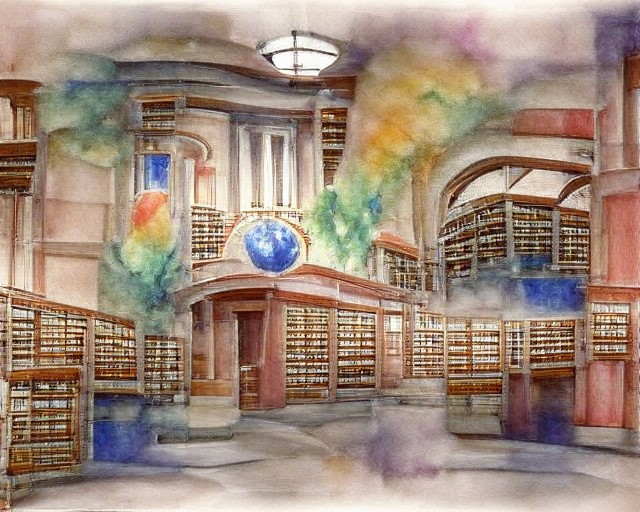}}\hfill
    \subfloat{\includegraphics[width=0.2\textwidth]{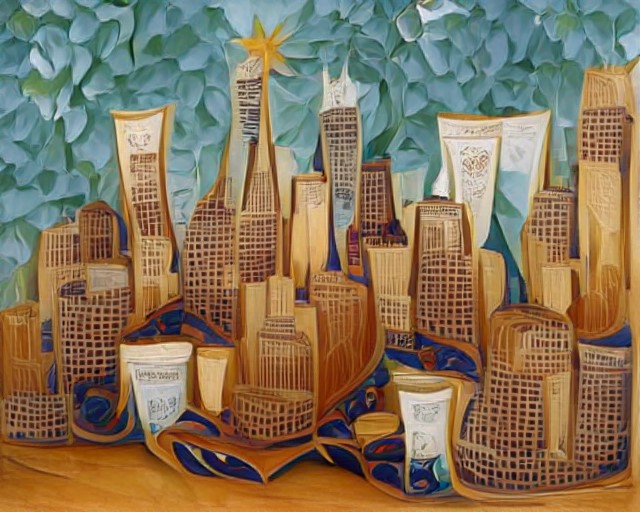}}\hfill
    \subfloat{\includegraphics[width=0.2\textwidth]{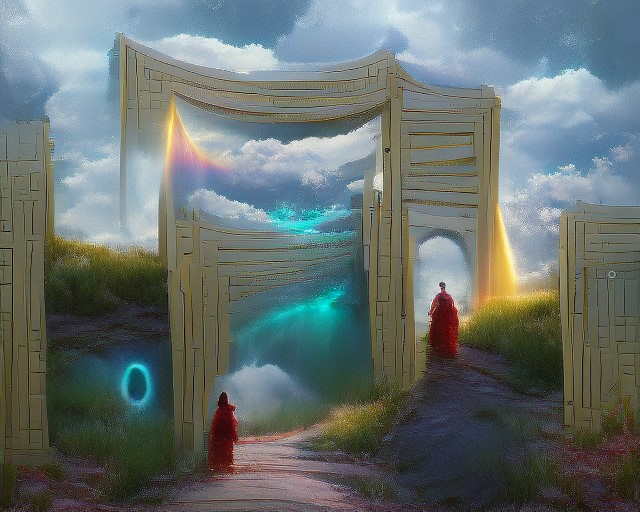}}\\
    \subfloat{\includegraphics[width=0.2\textwidth]{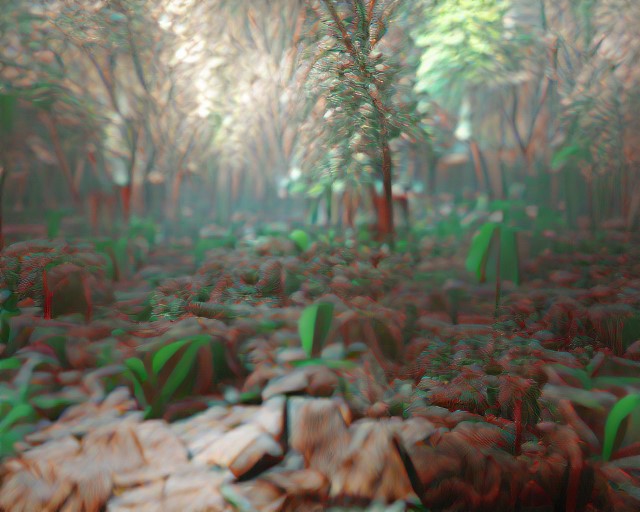}}\hfill
    \subfloat{\includegraphics[width=0.2\textwidth]{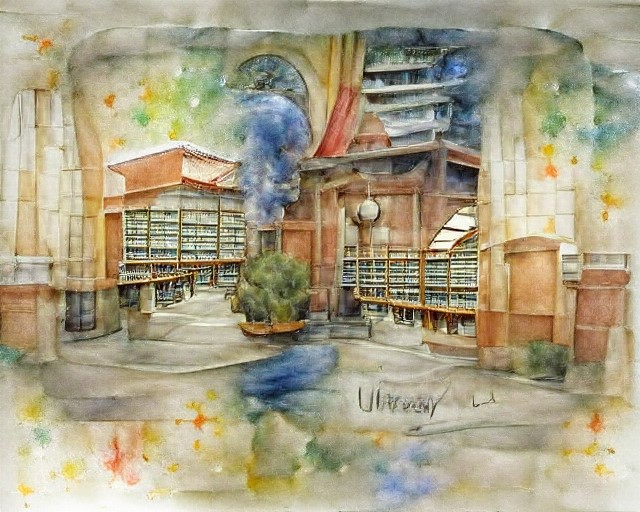}}\hfill
    \subfloat{\includegraphics[width=0.2\textwidth]{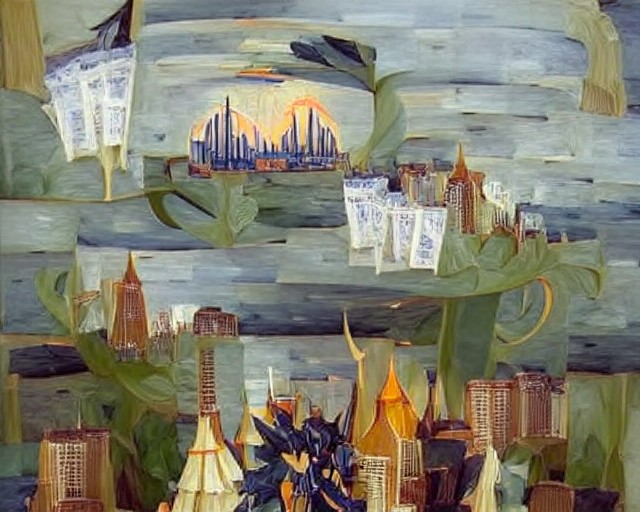}}\hfill
    \subfloat{\includegraphics[width=0.2\textwidth]{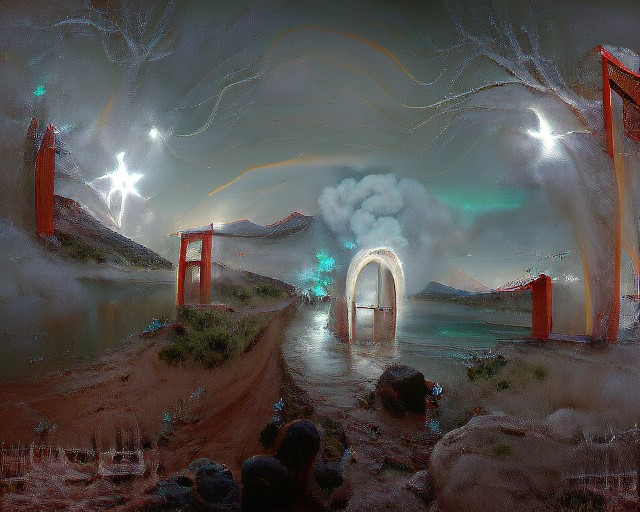}}\\
    \setcounter{subfigure}{0}
    \subfloat[a forest rendered in the Unreal Engine]{\includegraphics[width=0.2\textwidth]{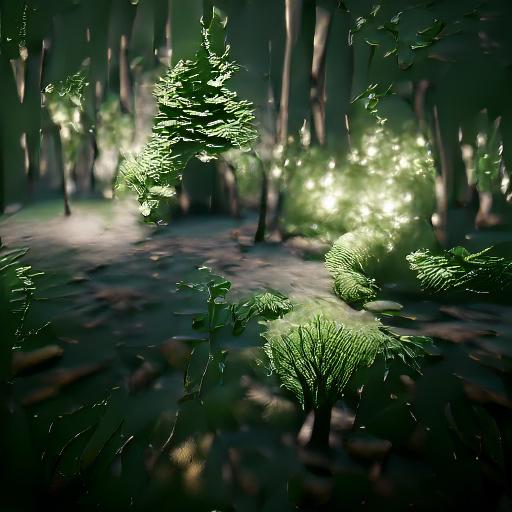}}\hfill
    \subfloat[a watercolor painting of the universal library]{\includegraphics[width=0.2\textwidth]{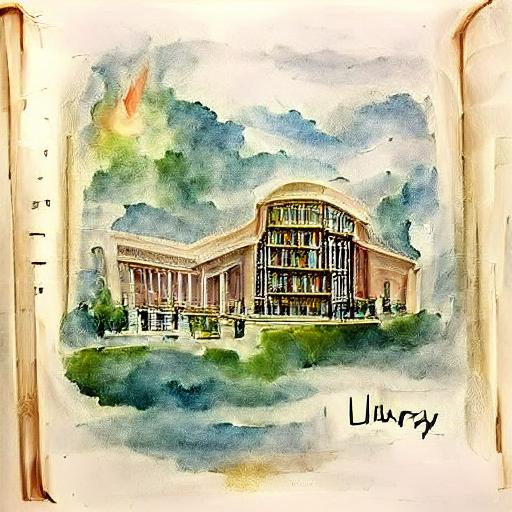}}\hfill
    \subfloat[An oil painting of The New York City Skyline by Natalia Goncharova]{\includegraphics[width=0.2\textwidth]{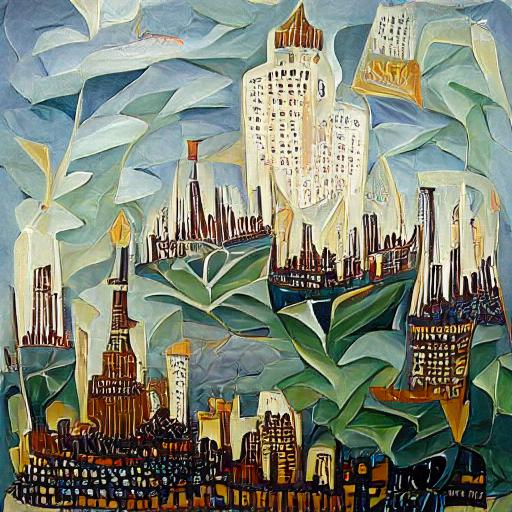}}\hfill
    \subfloat[the gateway between dreams trending on ArtStatio]{\includegraphics[width=0.2\textwidth]{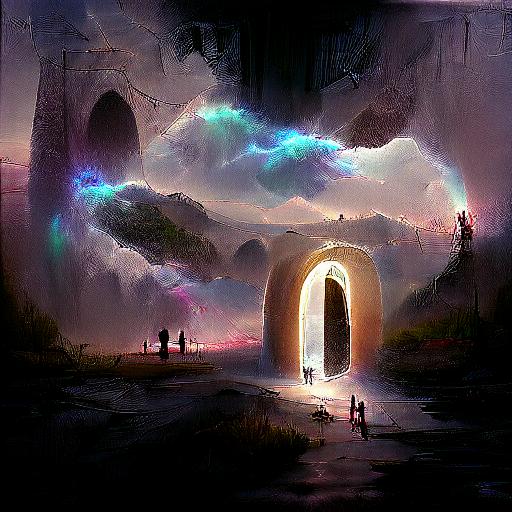}}\\
    \caption{Comparisons between codebook sampling (top), z-quantize (middle), and z-quantize with MSE regularization methods (bottom). Notice that z-quantize with regularization is able to produce finer details than alternative options, while non-regularized and codebook tend to yield  muddled images.}
    \label{fig:methods}
\end{figure}

Prior versions of this work used a methodology called Codebook Sampling, which optimizes a categorical distribution over a grid superimposed on the latent vectors. We found this approach was too slow for interactive use, and did leave considerable room for visual improvement. In \cref{fig:methods} we show the improvement in quality of our current approach compared to the prior. 
By adding a regularization term \cref{sec:regularize}, we obtain a methodology that is both faster and produces higher quality images than Codebook Sampling, giving the results presented in this paper. Our approach to regularization is incompatible with Codebook Sampling because of the tendency of Codebook Sampling to commit to particular codes early in training before the effects of regularization manifest.

\subsection{Number of Augmentations}

\begin{figure}[h]
\adjustbox{max width=\columnwidth}{%
\centering
\begin{tabular}{cccc}
 & no augmentation &  one augmentation  & all augmentations\\
\makecell{a child's drawing of a cathedral\\~\\ \textit{Random Affine Transformation}}
& \includegraphics[width=.25\linewidth,valign=m]{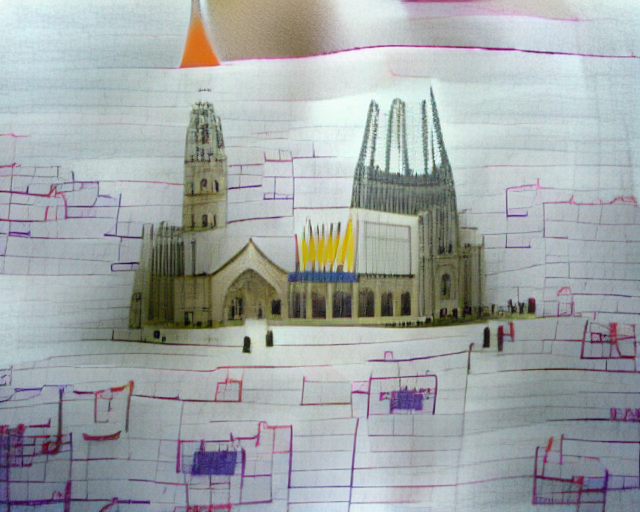}
& \includegraphics[width=.25\linewidth,valign=m]{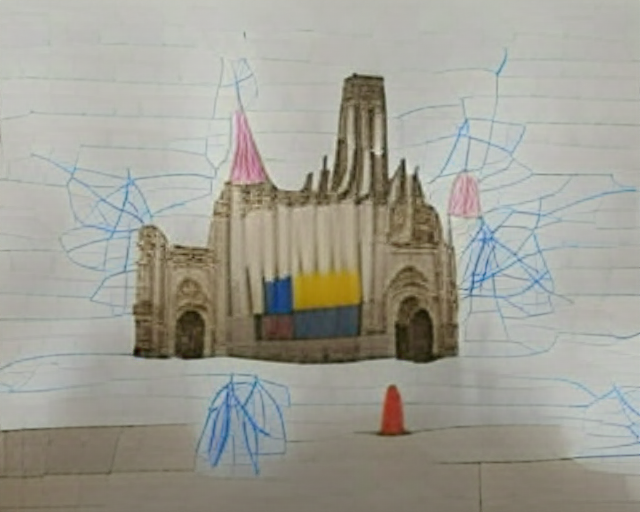}
& \includegraphics[width=.25\linewidth,valign=m]{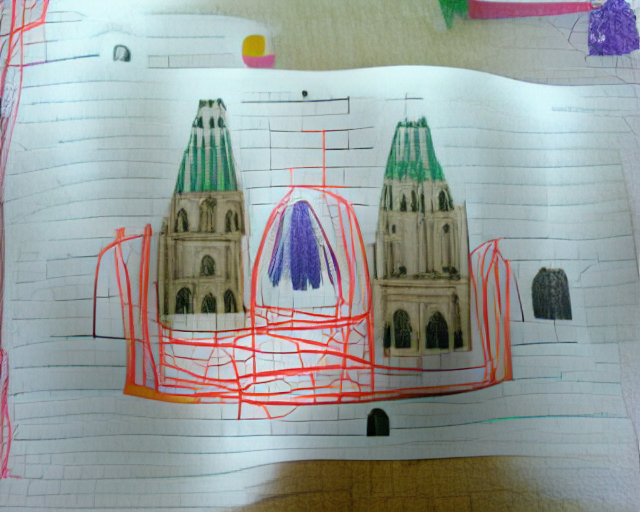}\\
\makecell{the universal library,\\rendered in the Unreal Engine\\~\\\textit{Random Perspective}}
& \includegraphics[width=.25\linewidth,valign=m]{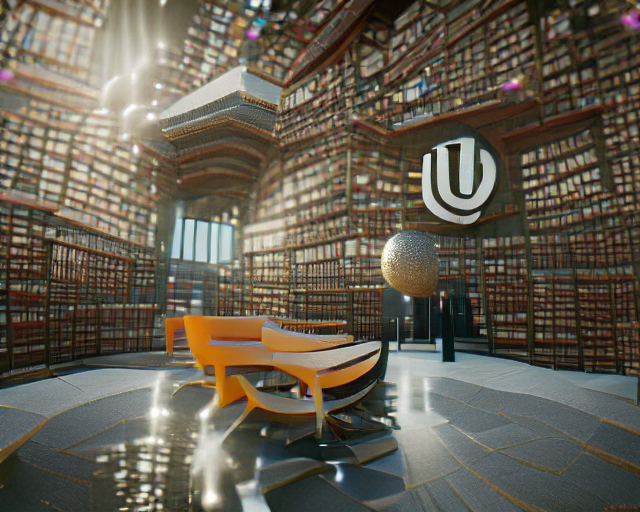}
& \includegraphics[width=.25\linewidth,valign=m]{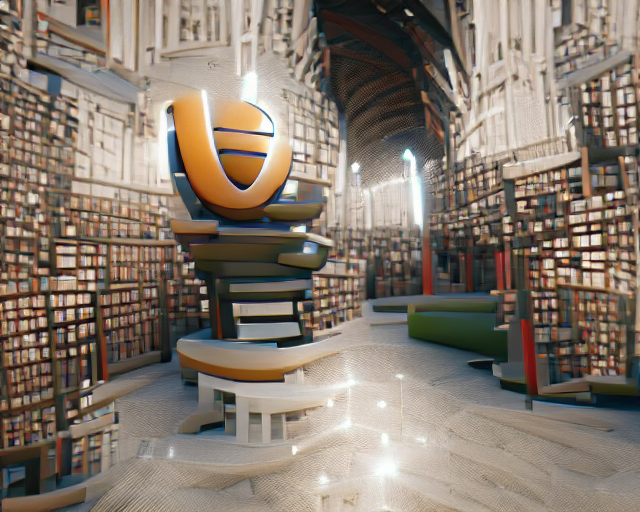}
& \includegraphics[width=.25\linewidth,valign=m]{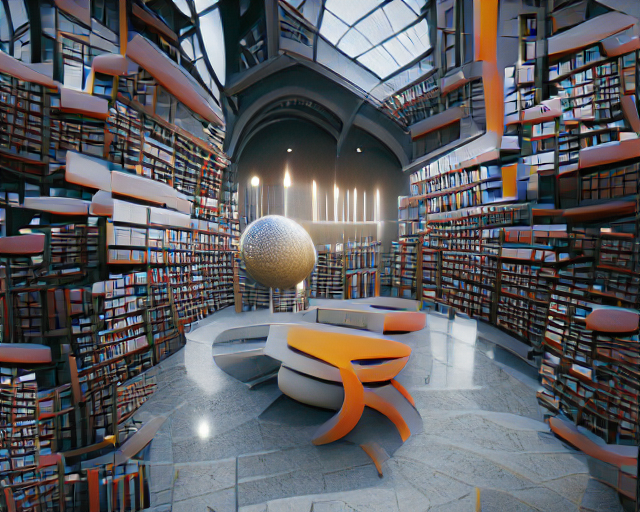}\\
\makecell{a cityscape at night,\\ rendered in the 8k resolution.\\~\\\textit{Random Noise}}
& \includegraphics[width=.25\linewidth,valign=m]{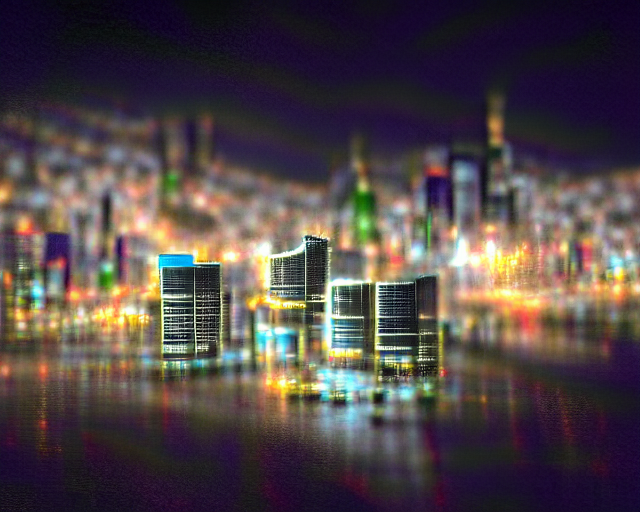}
& \includegraphics[width=.25\linewidth,valign=m]{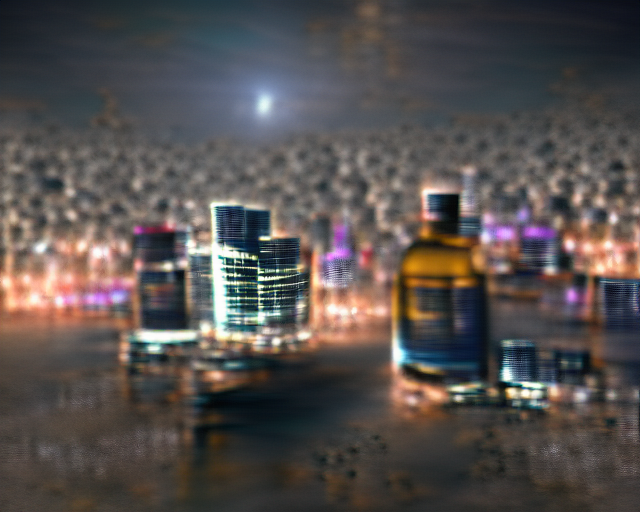}
& \includegraphics[width=.25\linewidth,valign=m]{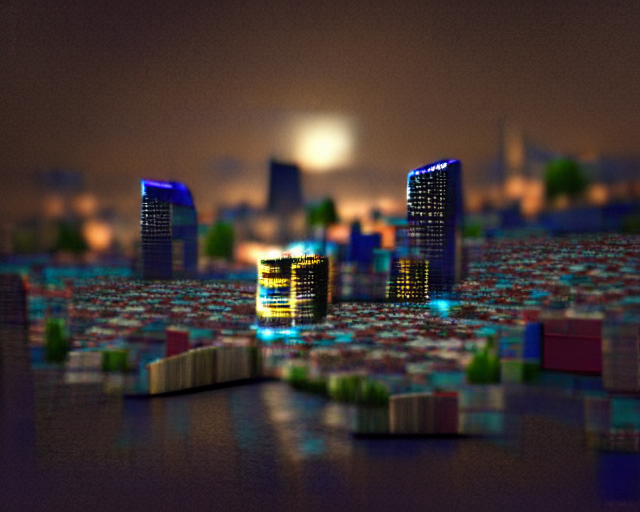}\\
\end{tabular}
}
\caption{Examples of the impacts of various augmentations. Left the right: unaugmented, only the named augmentation, all augmentations.} \label{fig:augmentations_num}
\end{figure}

In \cref{fig:augmentations_num} we ablate the number of augmentations to show their importance in consistent result quality. 

We find that the affine augmentation reduces clutter and unwanted duplicative generations, prospective augmentation improves the consistency of the 3D geometry, the noise augmentation improves the isolation of the foreground from the background.

\section{Resource Considerations}\label{sec:resource}

Our approach runs in $(935.2 \pm 20.4)\,\mathrm{s}$ on an NVIDIA Tesla K80 and $(229.5 \pm 26.2)\,\mathrm{s}$ on an NVIDIA GeForce RTX 2080 Ti (10 runs in each sample). This is approximately three times slower than minDALL-E and ten times slower than GLIDE (filtered) in our testing. Although we would like to analyze the trade-offs involved with the fact that both models required extensive pretraining none of the papers we compare to report their training requirements in enough detail to analyze the trade-offs brought about by this difference. We encourage the authors to release more information about their models so that more complete analysis can be done.

\subsection{Efficiency as a Value}

One of the goals of this research\footnote{The inclusion of this section is heavily inspired by \citet{birhane2022values}} is to increase the accessibility of AI image generative and editing tools. We have deliberately limited our approach to something that requires less than 11 GB of VRAM, so that it fits inside widely available commercial GPUs such as K80s. This GPU is particularly important from an accessibility standpoint, as it is the largest GPU that can be easily obtained using a free account on Google Colaboratory. The full generative process takes less than 3 minutes in a Google Colab notebook, making this a viable approach for anyone with access to the internet and a Google account.

Researchers with significantly more resources can obtain higher quality images using various augmentations left out of this paper, such as using an ensemble or additional auxiliary models to regularize the generations. While pushing the performance of our methodology to the maximum is a worthwhile endeavour, the fact that we can outperform the current state-of-the-art while running on freely available resources is something that we view as particularly worth highlighting. We leave determining the optimal framework with unbound resources to future work.

\subsection{Runtime Analysis}

DALL-E \citep{ramesh2021zero}, GLIDE \citep{nichol2021glide}, and Open-Edit \citep{liu2020open} all also incorporate image generators and joint text-image encoders into their architecture. Unlike our method however, they require computationally intensive training and finetuning. This invites the question of trade-offs between training and inference time. Unfortunately, none of the aforementioned papers report their training requirements in enough detail to estimate their training requirements. We are however able to estimate how long minDALL-E \citep{kakaobrain2021minDALL-E}, the current state-of-the-art DALL-E model, takes to train at $504$ V100-hours for the base model plus an additional $288$ V100-hours to finetune on ImageNet \citep{deng2009imagenet}. Through private communication with the authors, we were able to learn that GLIDE (filtered) required $400$ A100-days to train, which we approximate as $19,200$ V100-hours for ease of comparison.

\begin{table}[]
    \centering
    \begin{tabular}{lcccc}
        Model             & K80                & P100               & V100              & Training \\\toprule
        minDALL-E         & $216.0s \pm 07.6s$ & $60.0s \pm 5.5s$   & $016.3s \pm 2.7s$ & $792$ V100-hours\\
        GLIDE (filtered)  & $096.2s \pm 00.1s$ & $19.2s \pm 0.3s$   & $009.7s \pm 1.1s$ & $19,200$ V100-hours \\
        \VC{}             & $935.2s \pm 20.4s$ & $654.3s \pm 10.1s$ & $188.3s \pm 1.2s$ & $0$ V100-hours \\\bottomrule\\
    \end{tabular}
    \caption{Run-time of minDALL-E, GLIDE (filtered) and \VC{} on a variety of GPUs. Each cell shows the mean and standard deviation of a 10-run sample. We exclude Open-Edit due to a lack of information about its training costs. minDALL-E becomes cheaper than \VC{} after $858$ V100-hours have been expended while GLIDE (filtered) requires $20200$ V100-hours.}
    \label{tab:my_label}
\end{table}

On all hardwares evaluated, our model is substantially slower than both minDALL-E and GLIDE (filtered). However. In terms of trade-offs between training and inference on V100 GPUs, minDALL-E's total cost becomes cheaper than \VC{} at $\approx 15,800$ generations, while GLIDE(filtered) requires $\approx 384,000$. In terms of compute expended, minDALL-E becomes cheaper than \VC{} after $858$ V100-hours while GLIDE (filtered) requires $20200$ V100-hours. While cost and efficiency concerns depend significantly on individual contexts, the fact that GLIDE (filtered) only becomes as efficient as \VC{} efficient after tens of thousands of dollars of compute have been expended substantially limits researchers' ability to experiment with and iterate on the methodology. The same applies to minDALL-E, albeit with a price tag in the thousands rather than tens of thousands.

\section{Adoption of VQGAN-CLIP} \label{sec:useage}

A unique aspect of \VC{} has been its public development over the past year, which has resulted in an active community of users and real-world impact within and beyond classical computer vision. \citet{kwon2021clipstyler,frans2021clipdraw,chendiffvg+,liu2021fusedream,tian2021modern} create additional components (see \cref{sec:additional}) that they insert into our framework to improve performance in particular target domains. \citet{michel2021text2mesh,fei2021wenlan,nichol2021glide} evaluate their pretrained models by substituting them in for VQGAN or CLIP in our framework. \citet{avrahami2021blended,gu2021vector} use our framework with diffusion models in place of GANs.

Beyond computer vision, \citet{yang2021words} show that it is useful in the materials engineering design processes. \citet{wu2021wav2clip,jang2022music2video} builds on our work by using the framework to perform sound-guided image generation. In the domain of affective computing and HCI, \citet{galanos2021affectgan} has further found \VC{} able to elicit targeted emotions from viewers.

This last example helps explain the widespread commercial adoption of \VC{}, with over a dozen commercial apps built to provide it as a service and over 500 NFTs produced using our method sold\footnote{\url{https://twitter.com/quasimondo/status/1438601395087753217}}. A sampling of commercial websites using \VC{} include
NightCafe, Wombo Art, snowpixel.app, starryai.com, neuralblender.com, and hypnogram.xyz. Collectively, across these sites, \VC{} has been used over 10 million times, showing the veracity of our approach to handle unstructured and diverse user content.

\section{Conclusion} \label{sec:conclusion}

We have presented \VC{}, a method of generating and manipulating images based on only human written text prompts. The  quality of our model's generations have high visual fidelity and remain faithful to the textual prompt, outperforming prior approaches like DALL-E and GLIDE. The fidelity has been externally validated by commercial success and use by multiple companies.  Compared to the only comparable approach to text based image editing, \VC{} continues to produce higher quality visual images --- especially when the textual prompt and image content have low semantic similarity. Our ablation studies show the use of augmented images during the optimization process is a key factor to our method's success and that a code-book approach would sacrifice both speed and quality in our approach.

\subsection*{Acknowledgements}

We would like to acknowledge Ryan Murdock, who developed a very similar technique for combining VQGAN and CLIP simultaneously to us \citep{advadnoun2021tweet1,advadnoun2021tweet2} but did not release his approach. Ryan was invited to coauthor this paper and declined. We are also indebted to the thousands of people who have used and built upon our methods and given us feedback that has allowed us to continue to improve our techniques.

Many of the experiments in this paper were made possible by CoreWeave. CoreWeave also provides compute for a free public demo in the EleutherAI Discord server which enabled the methods described in this paper to be iterated on rapidly by a great many people. The bot that runs the demo was created by Discord user BoneAmputee.

\printbibliography

\appendix
\section{Author Contributions}

\paragraph{Katherine Crowson:} Originated the idea of combining \textsc{VQGAN} with \textsc{CLIP} and lead the development of the approach.

\paragraph{Stella Biderman:} Lead the experimentation and writing of the paper.

\paragraph{Daniel Kornis:} developed latent vector regularization and carried out experiments on Open-Edit.

\paragraph{Dashiell Stander:} Assisted with generative experiments and the writing of the paper.

\paragraph{Eric Hallahan:} Assisted with generative experiments and the writing of the paper.

\paragraph{Louis Castricato:} Developed semantic image editing and masking techniques.

\paragraph{Edward Raff:} Advised the project, proposed ablation experiments, and assisted in the writing of the paper.

\section{Additional Related Work}

\subsection{Image Generation and Editing}

While there is a significant literature on semantic image generation and editing beyond those cited in the main body of the paper, it is overwhelmingly work that imposes strong constraints on and assumptions about the inputs it work with. Closed domain work such as \citet{dong2017semantic,nam2018text,li2020manigan,patashnik2021styleclip} achieves much higher quality images, but comes at the cost of being afunctional outside the domain the model was trained on. Other authors simplify the problem by using non-textual auxiliary inputs \citep{shocher2020semantic,ntavelis2020sesame} or filtering the inputs to a small, pre-defined vocabulary \citep{hu2021muse}.

\subsection{CNN Visualisation and Interpretability}

Our approach is surprisingly similar to older work on visualizing and interpreting convolutional neural networks. Techniques for identifying what region(s) of an image lead to it being assigned a label by a classifier such as \citet{simonyan2014deep,selvaraju2017grad,yosinski2015understanding} operate on similar principles to our backpropagating of the CLIP loss, and DeepDream \citep{mordvintsev2015deepdream} uses the same core idea of iteratively updating an image to match a label.

\subsection{Building Multimodal Models Cheaply}

Due to the high cost of training large, state-of-the-art transformer models \citep{sharir2020cost,black2022gpt}, methodologies that allow one to build off of previously trained models cheaply are highly valuable. The bulk of the existing literature on this topic focuses on editing or updating existing (typically unimodal) models in both computer vision \citep{bau2020rewriting} and natural language processing \citep{de2021editing,mitchell2021fast,matena2021merging}.

Another thread of research aims to leverage prompting to efficiently produce multimodal models from unimodal ones. \citet{tsimpoukelli2021multimodal} shows that an input-dependent version of prompt tuning \citep{lester2021power} enables them to convert a 7 billion parameter language model into a multimodal model by only training 68 million of the parameters. Building on their success, \citet{Eichenberg2021MAGMAM} replace the prompt tuning with adapters \citep{houlsby2019parameter} and incorporate CLIP embeddings, enabling them to train a state-of-the-art visual question answering transformer at a fraction of the cost of doing so from scratch. Our work continues in this theme by removing the training entirely, at the cost of a modest increase in generation time.

\section{Observations of Public Use}\label{app:public-use}

As our models have been public for nearly a year at time-of-writing, we have had the opportunity to study how people use \VC{} in the wild. While a systematic study is far outside the scope of this paper, we summarize our main observations in the hopes of guiding future human-computer interaction investigations. These observations are largely informed by casual conversations and interviews with users of our methodology, and by \textit{in situ} observation of people interacting with the model. For further discussion of the HCI and sociological influences on generative modeling, we direct an interested reader to \citet{snell2020alien,underwood2021mapping,ali2021telling}.

\paragraph{Human-AI Co-Creation} The overwhelming feedback we have gotten from users is that they view \VC{} not as an AI working in a vacuum but as an AI and a human working together to generate art. Users have also created new (and unintended) ways of working with \VC{} that increase their direct agency over the art created. The most prominent example of this is that many users halt generation early to edit the partial generations directly, before restarting generation from their newly edited image. In this way users can remove artifacts, discourage undesirable components of an image, and try alternative paths through the optimization landscape.

\paragraph{Notebook-based development} The overwhelming majority of people who have picked up and iterated on the methodologies described in this paper use notebooks such as Jupyter Notebook and Google Colab to do their work, instead of traditional repositories and scripts. Based on our conversations with users, the cause of this appears to be two-fold: widespread popularity among people who do not identify as computer scientists or developers and a lack of the necessary computational resources to run the model locally. Similar considerations drive the popularity of websites that offer \VC{}-as-a-service and the bot in our Discord server.

\paragraph{Iterated prompting} While the typical methodology in machine learning for improving results from an algorithm focus on improvements to the training data or algorithm, our users have exhibited a strong preference for modifying the \textit{prompts} they provide instead. We view this as exemplifying ``natural language as an API,'' \citep{sayers2021dawn} wherein the actual mapping between the images and the text is less important than the fact that text is a natural medium for human interaction. The end-goal is to obtain desirable images \textit{using whatever text input is necessary}.

\section{Additional Generations}\label{app:generations}
The following is a selection of artwork generated using our technique by people other than the authors of this paper, often with highly complex prompting. We include them both to emphasize the diversity of artistic styles \VC{} is capable of, as well as its widespread adoption. These images have served as the covers of academic journals, sold for thousands of dollars, formed the basis for physical paintings, and been displayed in art galleries.

\begin{figure}[ht]
\centering
\subfloat{\includegraphics[width=0.2\textwidth,height=0.2\textwidth]{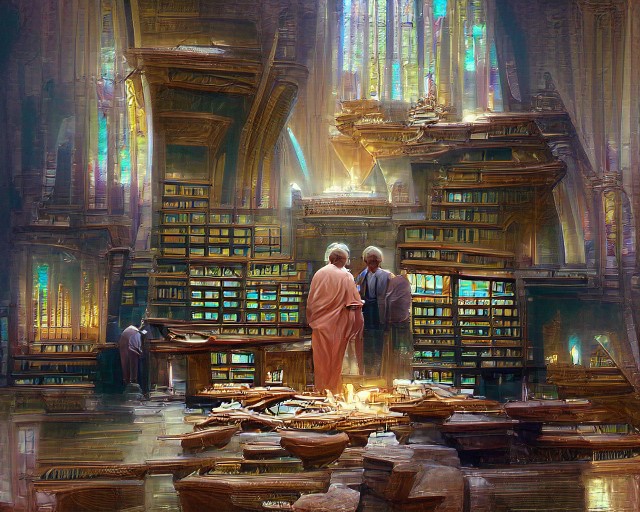}}\hfill
\subfloat{\includegraphics[width=0.2\textwidth,height=0.2\textwidth]{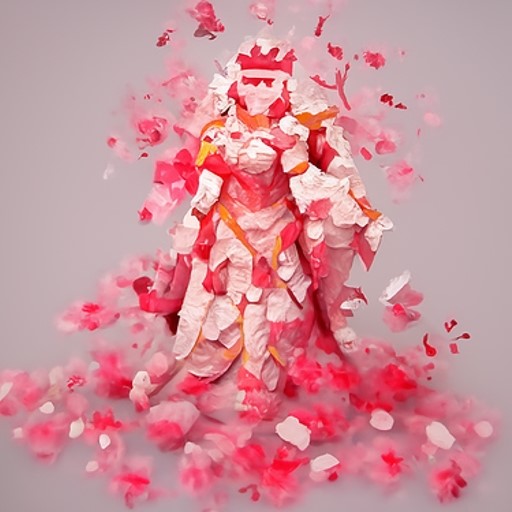}}\hfill
\subfloat{\includegraphics[width=0.2\textwidth,height=0.2\textwidth]{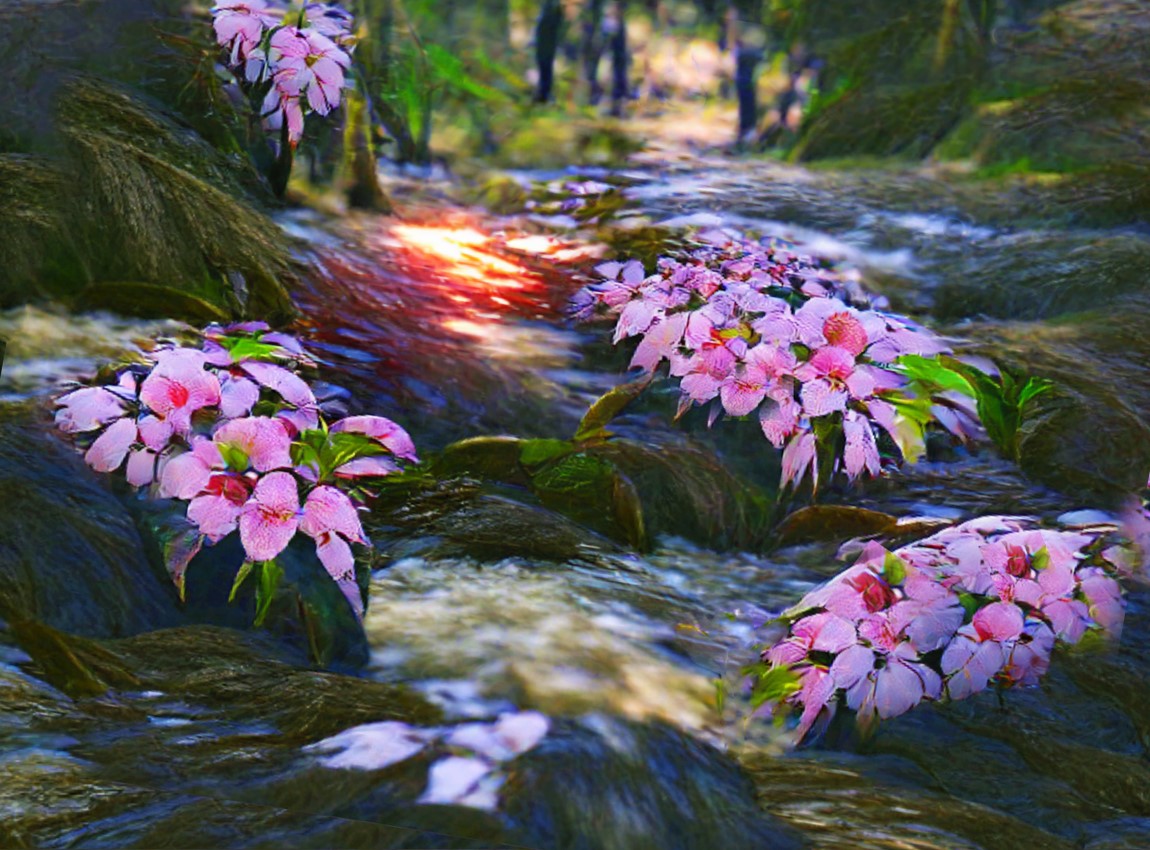}}\hfill
\subfloat{\includegraphics[width=0.2\textwidth,height=0.2\textwidth]{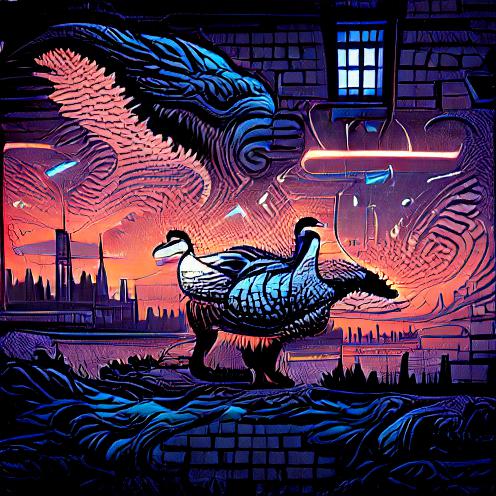}}\hfill\\
\subfloat{\includegraphics[width=0.2\textwidth,height=0.2\textwidth]{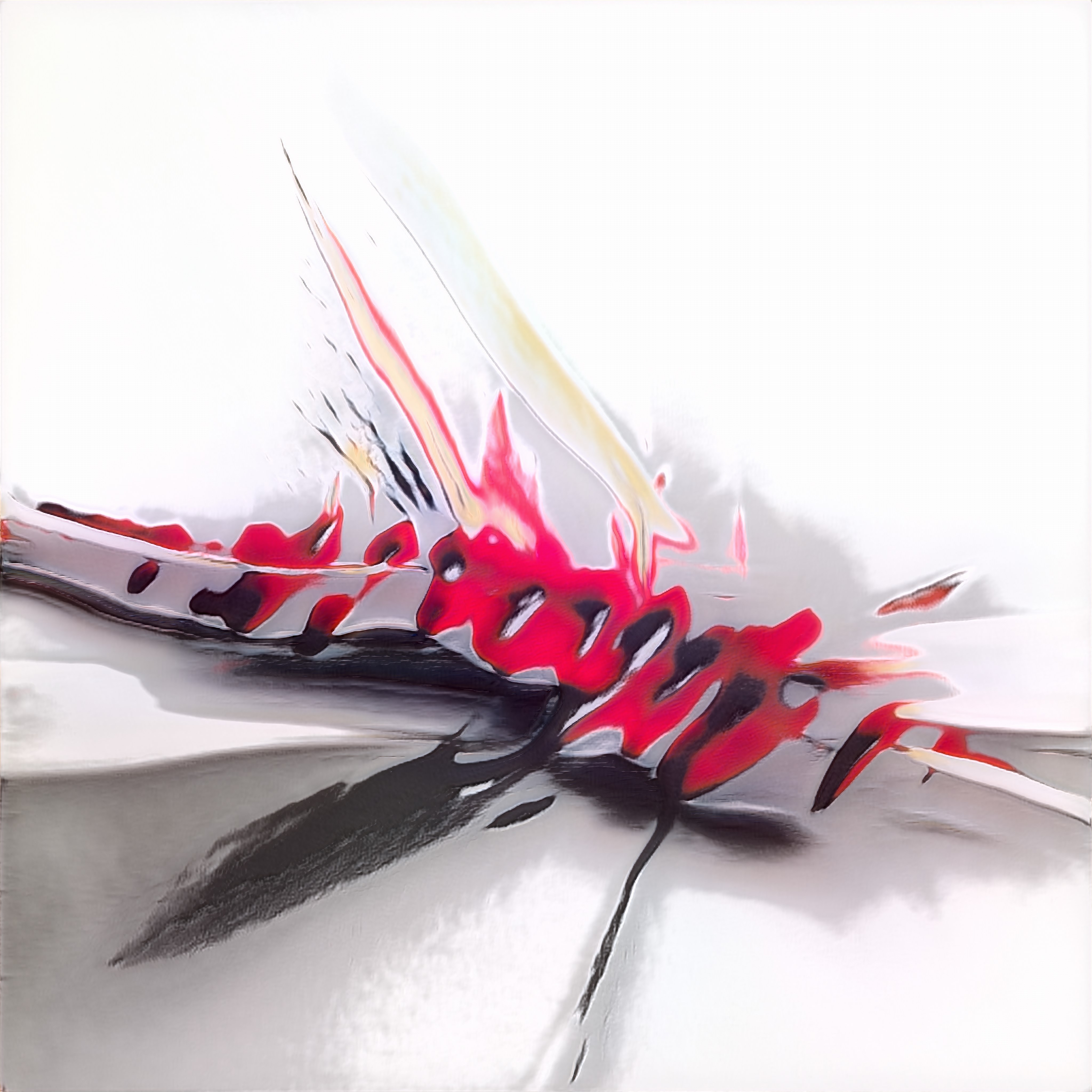}}\hfill
\subfloat{\includegraphics[width=0.2\textwidth,height=0.2\textwidth]{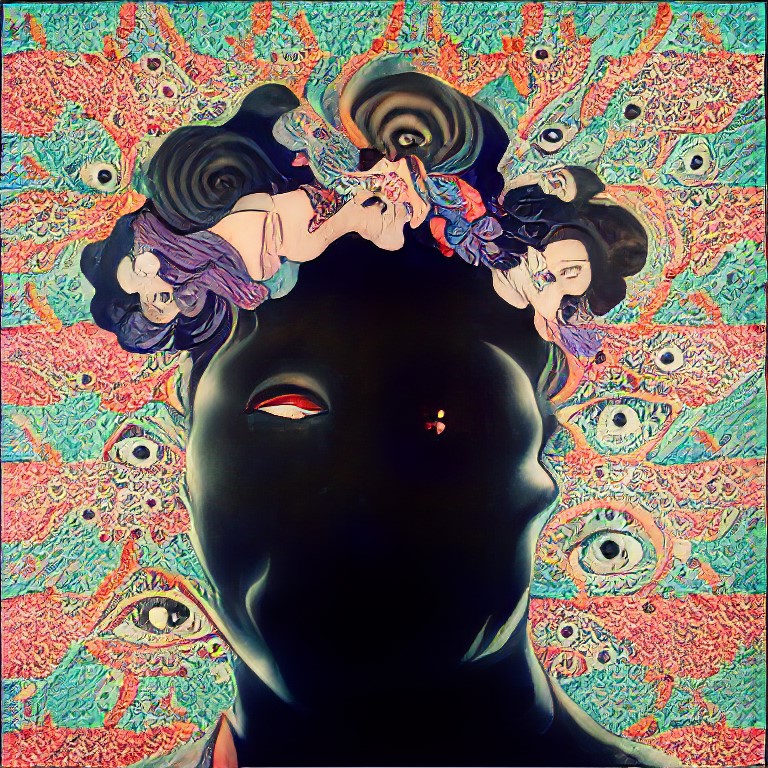}}\hfill
\subfloat{\includegraphics[width=0.2\textwidth,height=0.2\textwidth]{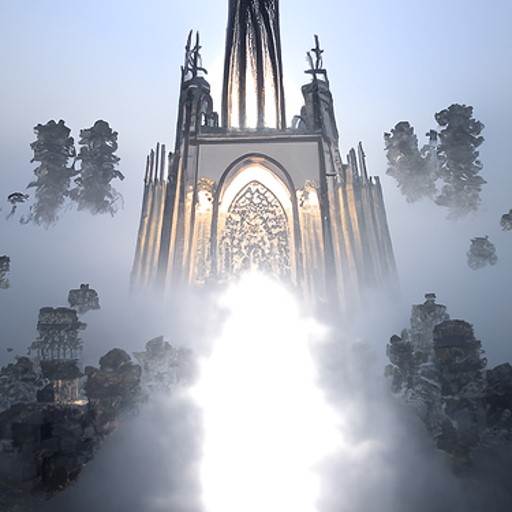}}\hfill
\subfloat{\includegraphics[width=0.2\textwidth,height=0.2\textwidth]{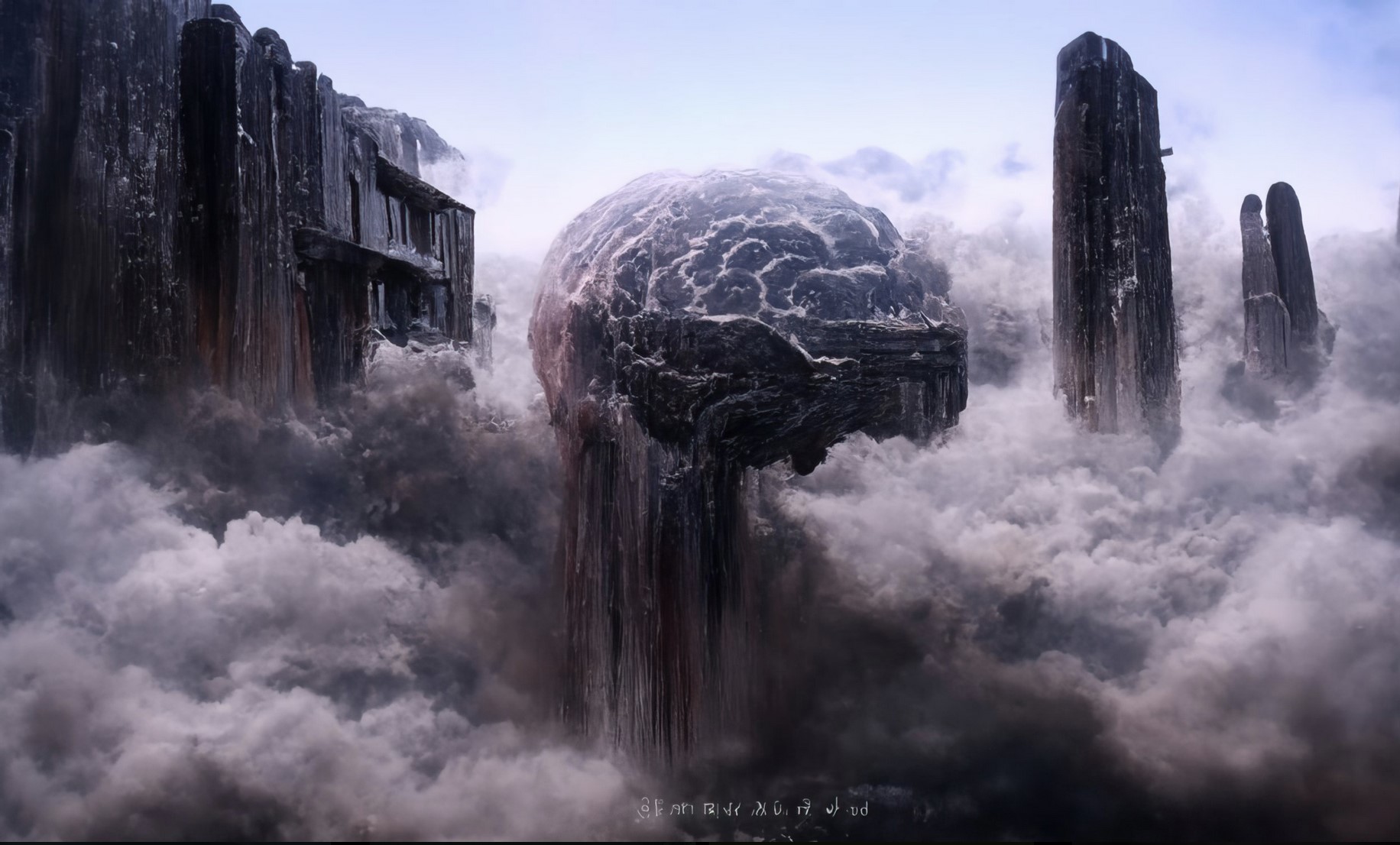}}\hfill\\
\subfloat{\includegraphics[width=0.2\textwidth,height=0.2\textwidth]{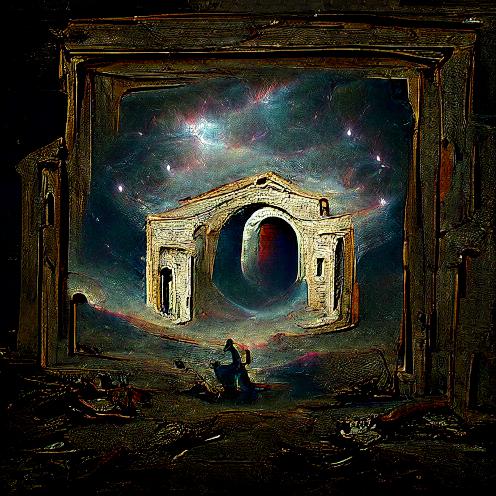}}\hfill
\subfloat{\includegraphics[width=0.2\textwidth,height=0.2\textwidth]{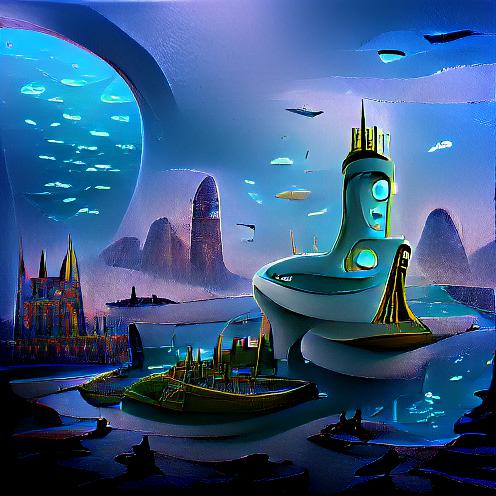}}\hfill
\subfloat{\includegraphics[width=0.2\textwidth,height=0.2\textwidth]{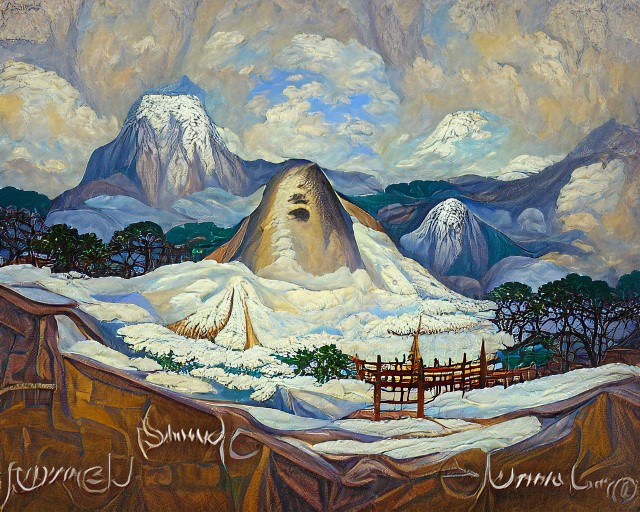}}\hfill
\subfloat{\includegraphics[width=0.2\textwidth,height=0.2\textwidth]{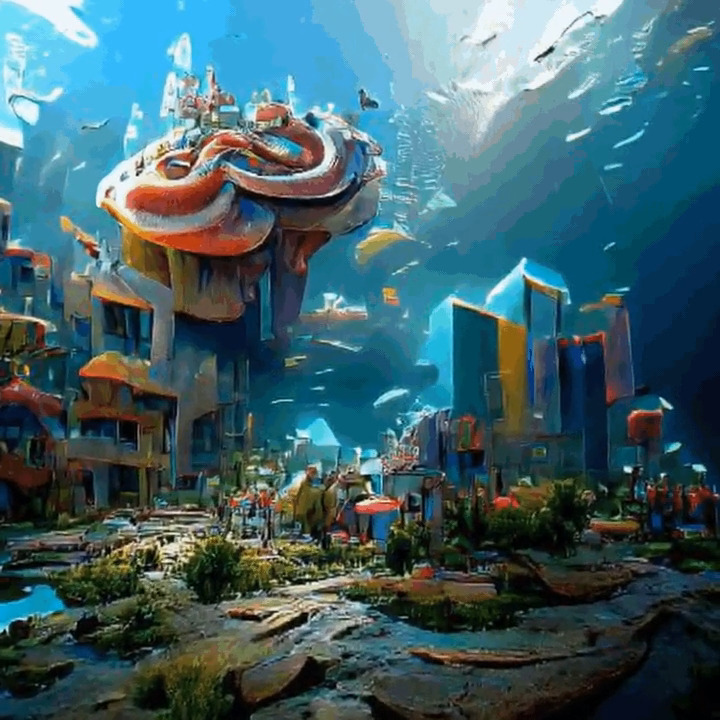}}\hfill\\
\subfloat{\includegraphics[width=0.2\textwidth,height=0.2\textwidth]{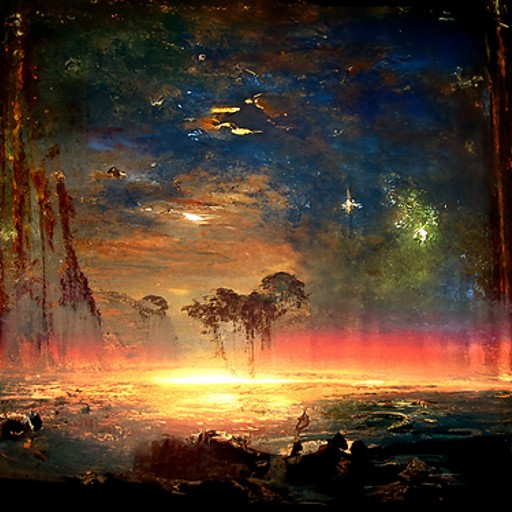}}\hfill
\subfloat{\includegraphics[width=0.2\textwidth,height=0.2\textwidth]{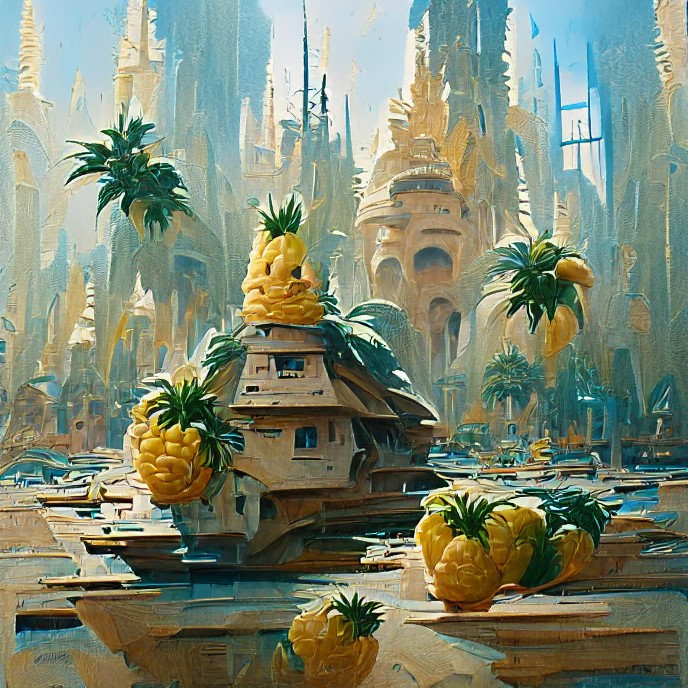}}\hfill
\subfloat{\includegraphics[width=0.2\textwidth,height=0.2\textwidth]{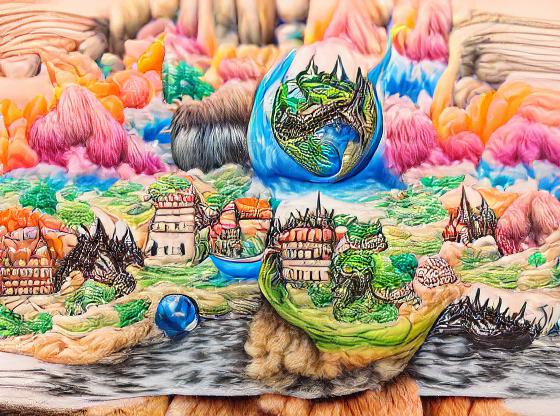}}\hfill
\subfloat{\includegraphics[width=0.2\textwidth,height=0.2\textwidth]{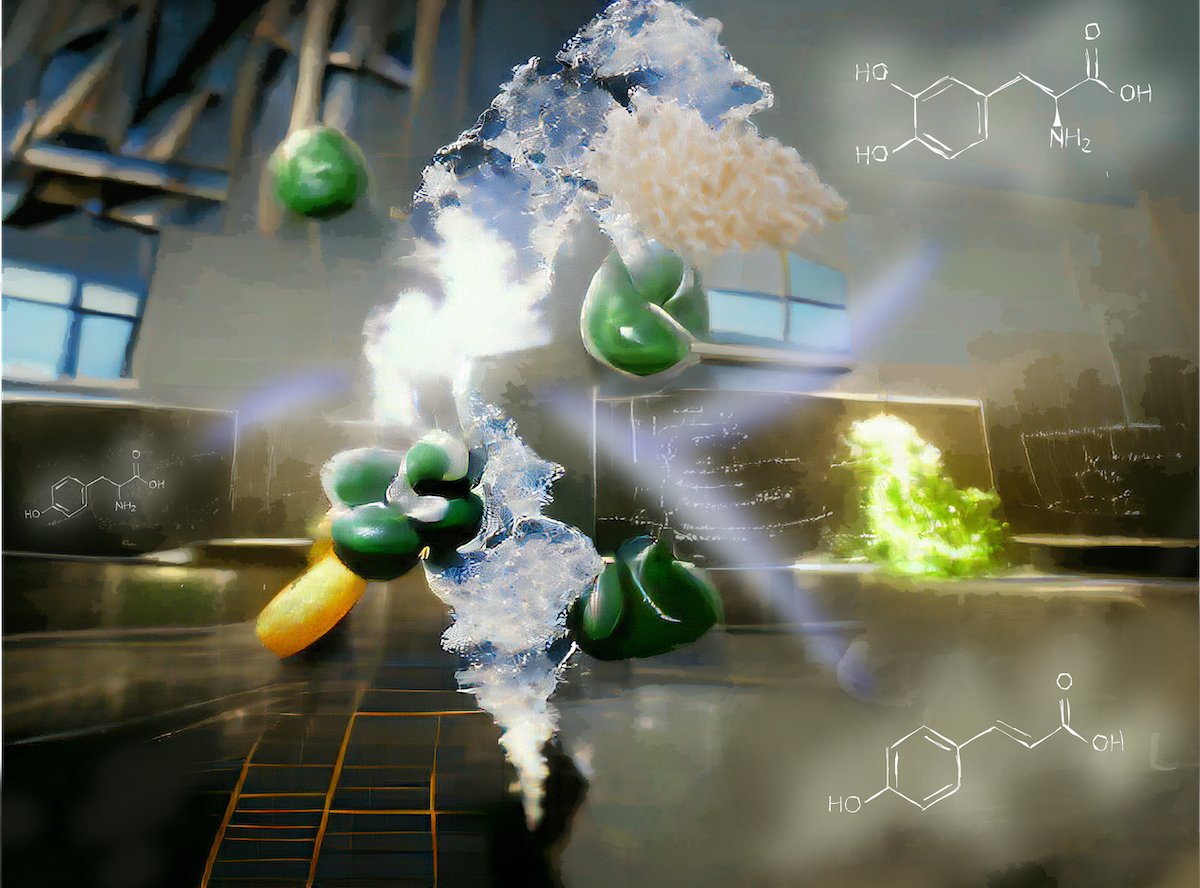}}\\
\subfloat{\includegraphics[width=0.2\textwidth,height=0.2\textwidth]{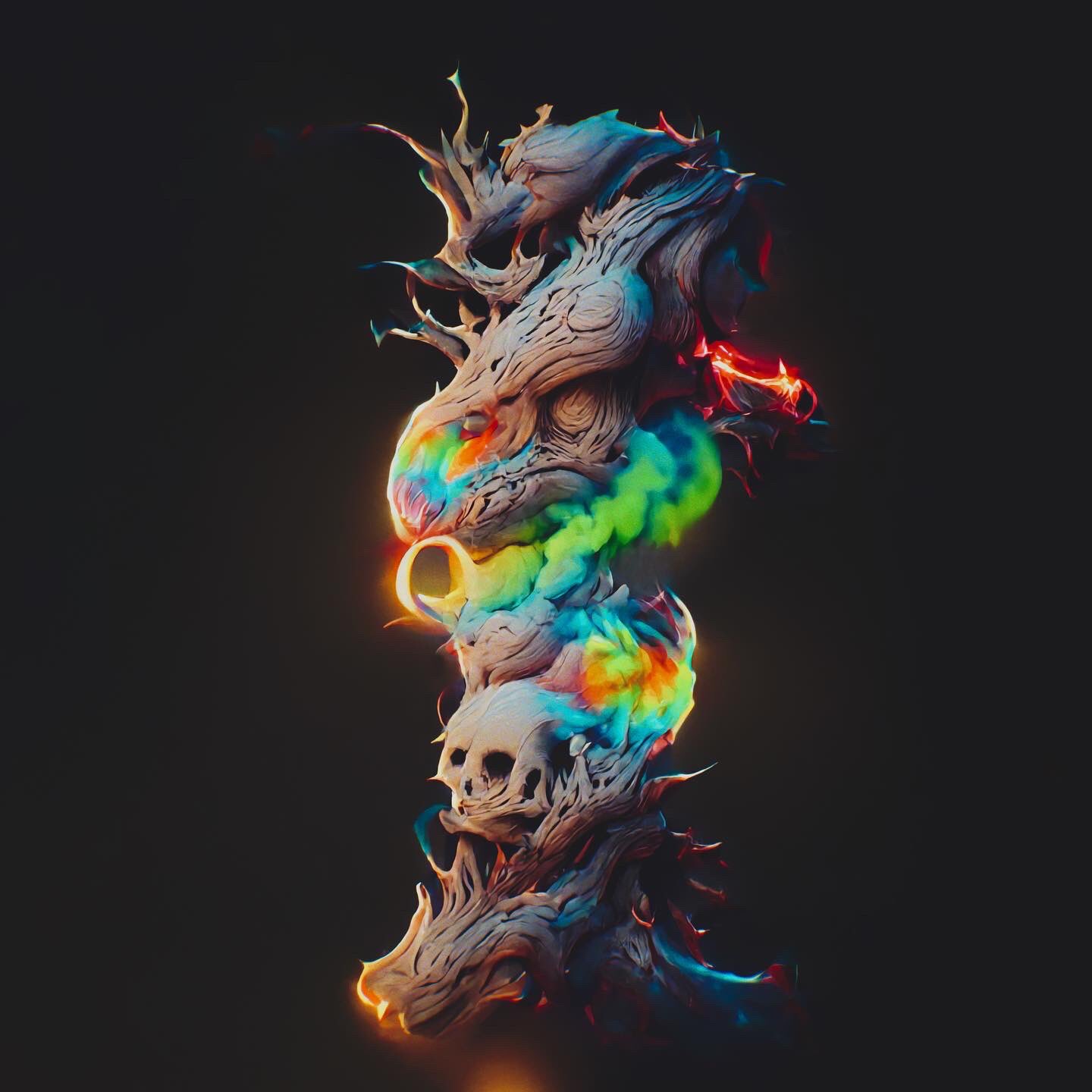}}\hfill
\subfloat{\includegraphics[width=0.2\textwidth,height=0.2\textwidth]{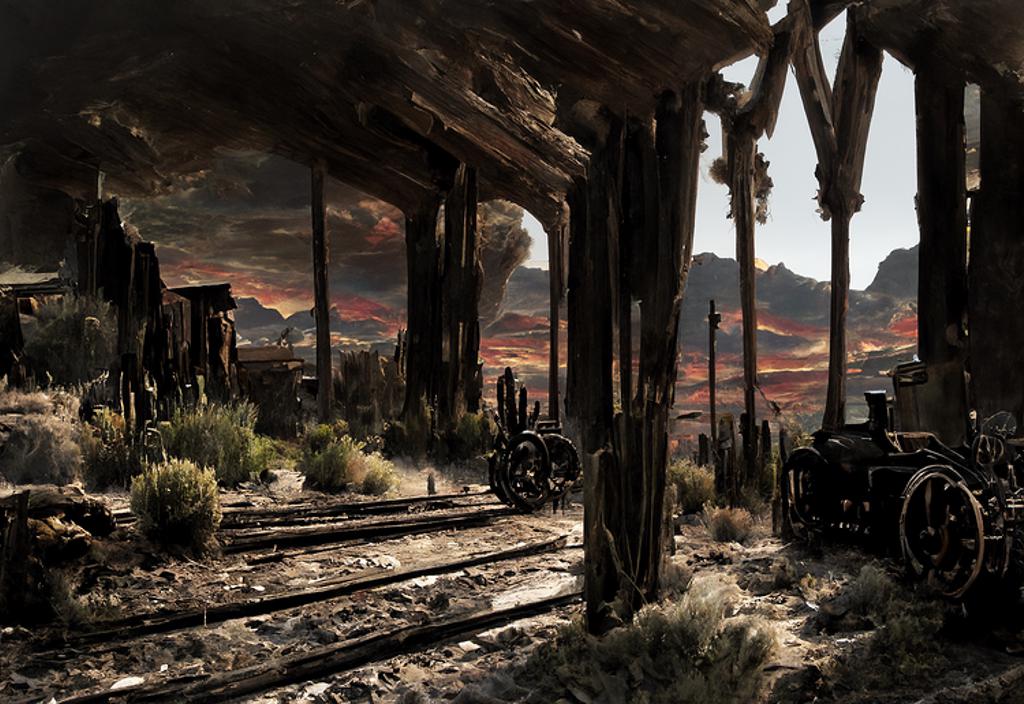}}\hfill
\subfloat{\includegraphics[width=0.2\textwidth,height=0.2\textwidth]{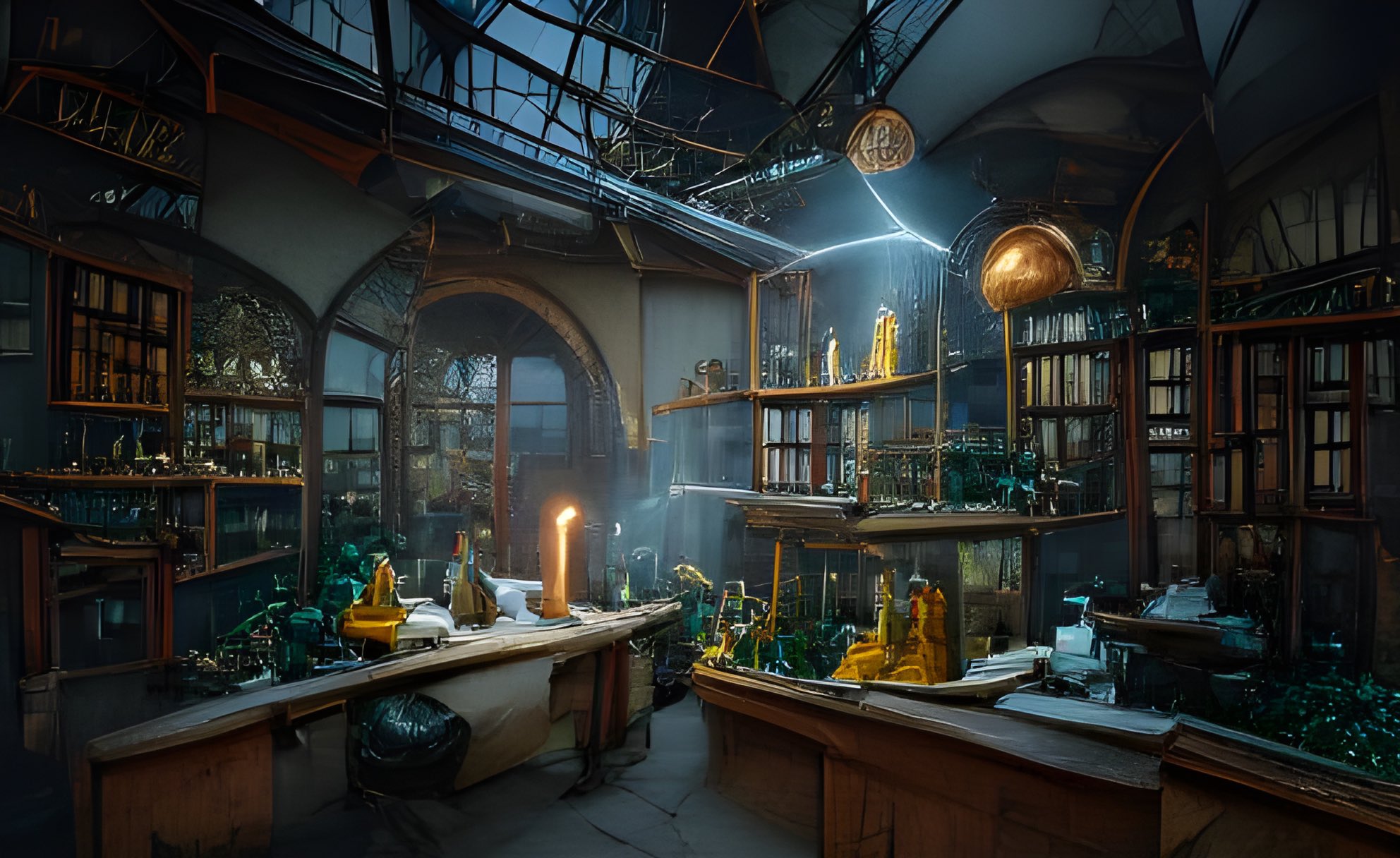}}\hfill
\subfloat{\includegraphics[width=0.2\textwidth,height=0.2\textwidth]{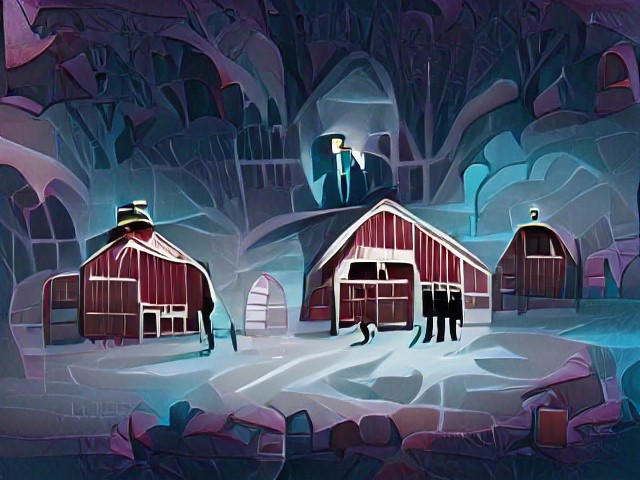}}
\end{figure}
\section{Additional Comparisons}\label{app:comparisons}

The following images showcase the minDALL-E \citep{kakaobrain2021minDALL-E} and GLIDE (filtered) \citep{nichol2021glide} generations with the prompts found in Figure 3.
\begin{figure}[h]
    \centering
    \subfloat[Oil painting of a candy dish of glass candies, mints, and other assorted sweets]{\includegraphics[width=0.25\textwidth]{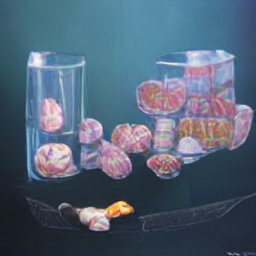}}\hfill
    \subfloat[A colored pencil drawing of a waterfall]{\includegraphics[width=0.25\textwidth]{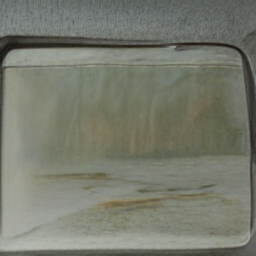}}\hfill
    \subfloat[A fantasy painting of a city in a deep valley by Ivan Aivazovsky]{\includegraphics[width=0.25\textwidth]{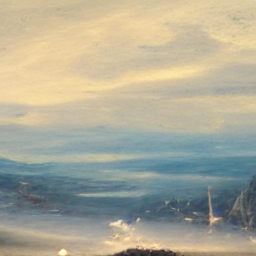}}\\
    \subfloat[A beautiful painting of a building in a serene landscape ]{\includegraphics[width=0.25\textwidth]{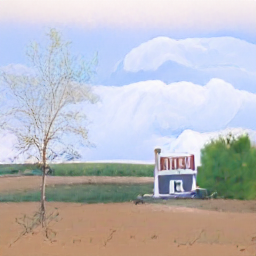}}\hfill
    \subfloat[sketch of a 3D printer by Leonardo da Vinci]{\includegraphics[width=0.25\textwidth]{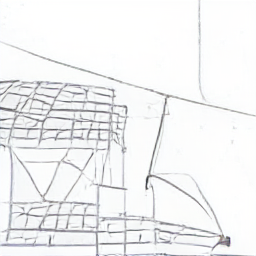}}\hfill
    \subfloat[an autogyro flying car, trending on artstation]{\includegraphics[width=0.25\textwidth]{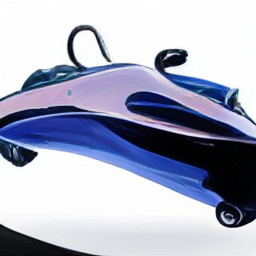}}\\
    \subfloat[an astronaut in the style of van Gogh]{\includegraphics[width=0.25\textwidth]{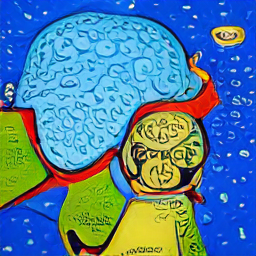}}\hfill
    \subfloat[Baba Yaga's house + fantasy art]{\includegraphics[width=0.25\textwidth]{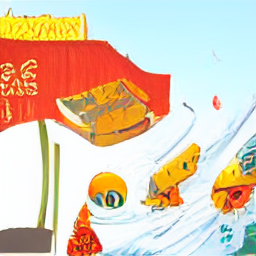}}\hfill
    \subfloat[pickled eggs, tempera on wood]{\includegraphics[width=0.25\textwidth]{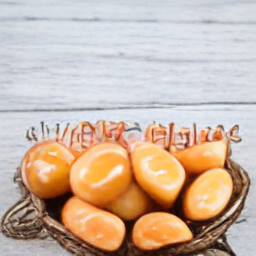}}\\\
    \subfloat[effervescent hope]{\includegraphics[width=0.25\textwidth]{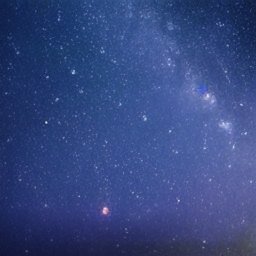}}\hfill
    \subfloat[the Tower of Babel by J.M.W. Turner]{\includegraphics[width=0.25\textwidth]{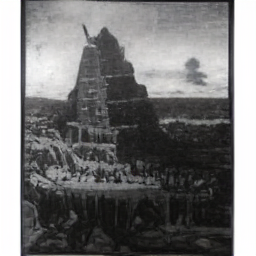}}\hfill
    \subfloat[a futuristic city in synthwave style]{\includegraphics[width=0.25\textwidth]{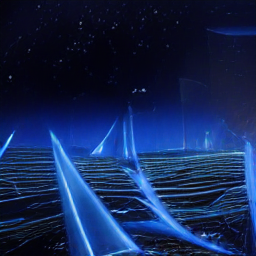}}
    \caption{Example minDALL-E generations and their text prompts.}
    \label{fig:example-gen-min}
\end{figure}

\begin{figure}
    \centering
    \subfloat[Oil painting of a candy dish of glass candies, mints, and other assorted sweets]{\includegraphics[width=0.25\textwidth]{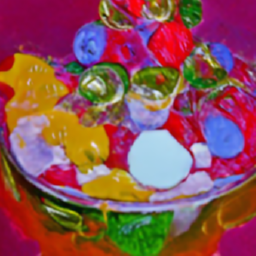}}\hfill
    \subfloat[A colored pencil drawing of a waterfall]{\includegraphics[width=0.25\textwidth]{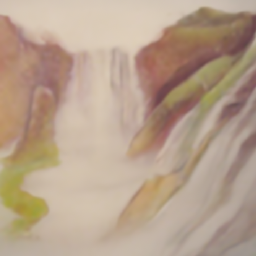}}\hfill
    \subfloat[A fantasy painting of a city in a deep valley by Ivan Aivazovsky]{\includegraphics[width=0.25\textwidth]{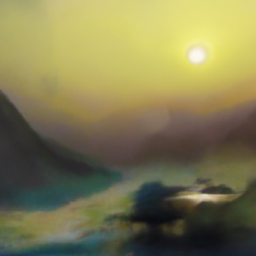}}\\
    \subfloat[A beautiful painting of a building in a serene landscape ]{\includegraphics[width=0.25\textwidth]{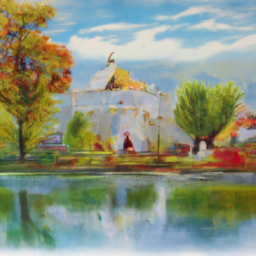}}\hfill
    \subfloat[sketch of a 3D printer by Leonardo da Vinci]{\includegraphics[width=0.25\textwidth]{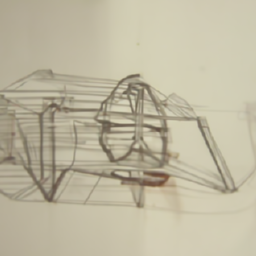}}\hfill
    \subfloat[an autogyro flying car, trending on artstation]{\includegraphics[width=0.25\textwidth]{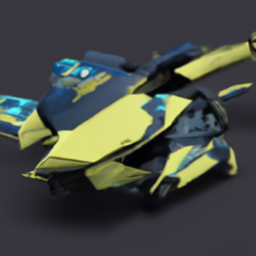}}\\
    \subfloat[an astronaut in the style of van Gogh]{\includegraphics[width=0.25\textwidth]{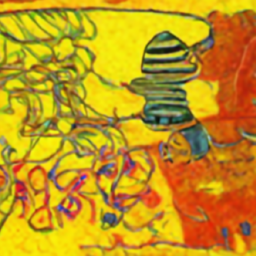}}\hfill
    \subfloat[Baba Yaga's house + fantasy art]{\includegraphics[width=0.25\textwidth]{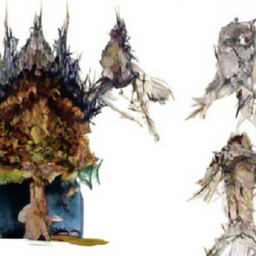}}\hfill
    \subfloat[pickled eggs, tempera on wood]{\includegraphics[width=0.25\textwidth]{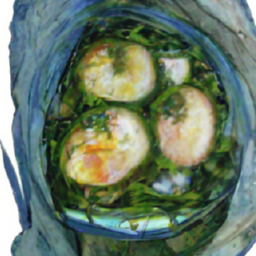}}\\\
    \subfloat[effervescent hope]{\includegraphics[width=0.25\textwidth]{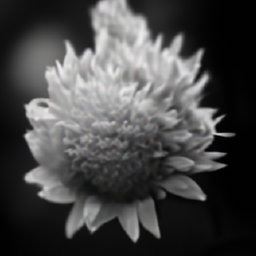}}\hfill
    \subfloat[the Tower of Babel by J.M.W. Turner]{\includegraphics[width=0.25\textwidth]{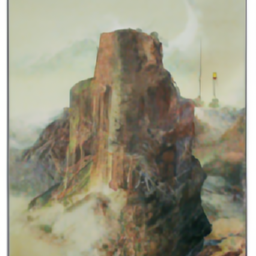}}\hfill
    \subfloat[a futuristic city in synthwave style]{\includegraphics[width=0.25\textwidth]{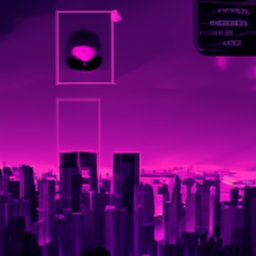}}
    \caption{Example GLIDE (CLIP-Guided) generations and their text prompts.}
    \label{fig:example-gen-clip}
\end{figure}

\begin{figure}
    \centering
    \subfloat[Oil painting of a candy dish of glass candies, mints, and other assorted sweets]{\includegraphics[width=0.25\textwidth]{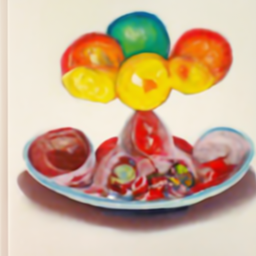}}\hfill
    \subfloat[A colored pencil drawing of a waterfall]{\includegraphics[width=0.25\textwidth]{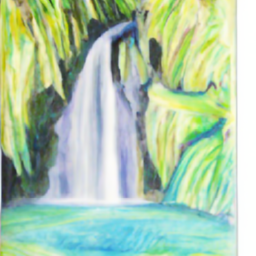}}\hfill
    \subfloat[A fantasy painting of a city in a deep valley by Ivan Aivazovsky]{\includegraphics[width=0.25\textwidth]{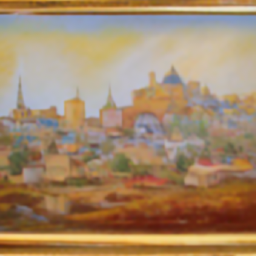}}\\
    \subfloat[A beautiful painting of a building in a serene landscape ]{\includegraphics[width=0.25\textwidth]{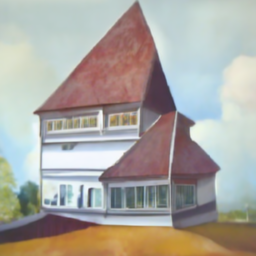}}\hfill
    \subfloat[sketch of a 3D printer by Leonardo da Vinci]{\includegraphics[width=0.25\textwidth]{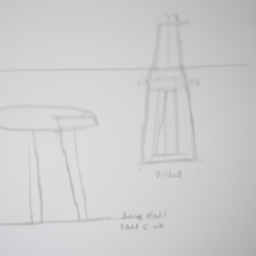}}\hfill
    \subfloat[an autogyro flying car, trending on artstation]{\includegraphics[width=0.25\textwidth]{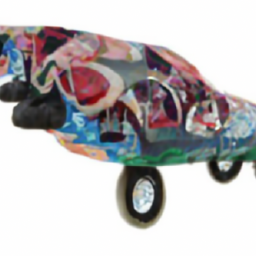}}\\
    \subfloat[an astronaut in the style of van Gogh]{\includegraphics[width=0.25\textwidth]{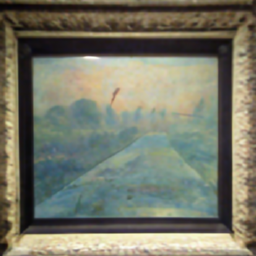}}\hfill
    \subfloat[Baba Yaga's house + fantasy art]{\includegraphics[width=0.25\textwidth]{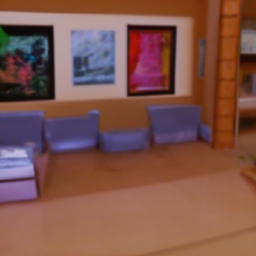}}\hfill
    \subfloat[pickled eggs, tempera on wood]{\includegraphics[width=0.25\textwidth]{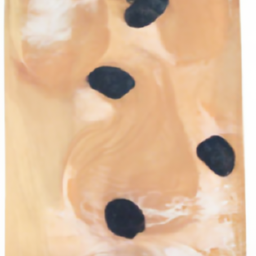}}\\\
    \subfloat[effervescent hope]{\includegraphics[width=0.25\textwidth]{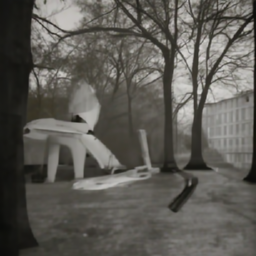}}\hfill
    \subfloat[the Tower of Babel by J.M.W. Turner]{\includegraphics[width=0.25\textwidth]{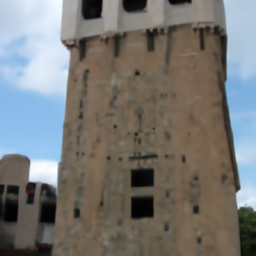}}\hfill
    \subfloat[a futuristic city in synthwave style]{\includegraphics[width=0.25\textwidth]{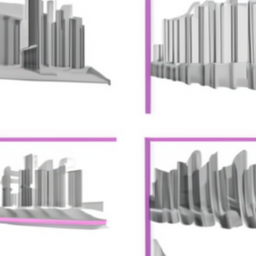}}
    \caption{Example GLIDE (CF-Guided) generations and their text prompts.}
    \label{fig:example-gen-glide}
\end{figure}

\begin{figure}
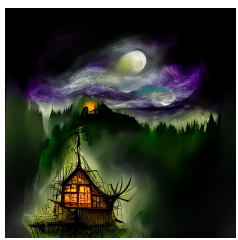
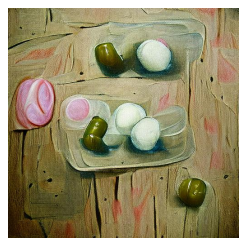

    \centering
    \subfloat[Oil painting of a candy dish of glass candies, mints, and other assorted sweets]{\includegraphics[width=0.25\textwidth]{VQGAN-CLIP/generation/candy.jpg}}\hfill
    \subfloat[A colored pencil drawing of a waterfall]{\includegraphics[width=0.25\textwidth]{VQGAN-CLIP/generation/waterfall.jpg}}\hfill
    \subfloat[A fantasy painting of a city in a deep valley by Ivan Aivazovsky]{\includegraphics[width=0.25\textwidth]{VQGAN-CLIP/generation/A_fantasy_painting_of_a_city_in_a_deep_valley_by_Ivan_Aivazovsky.jpg}}\\
    \subfloat[A beautiful painting of a building in a serene landscape ]{\includegraphics[width=0.25\textwidth]{VQGAN-CLIP/generation/serene.jpg}}\hfill
    \subfloat[sketch of a 3D printer by Leonardo da Vinci]{\includegraphics[width=0.25\textwidth]{VQGAN-CLIP/generation/3D.jpg}}\hfill
    \subfloat[an autogyro flying car, trending on artstation]{\includegraphics[width=0.25\textwidth]{VQGAN-CLIP/generation/autogyro.jpg}}\\
    \subfloat[an astronaut in the style of van Gogh]{\includegraphics[width=0.25\textwidth]{VQGAN-CLIP/generation/astronaut.jpg}}\hfill
    \subfloat[Baba Yaga's house + fantasy art]{\includegraphics[width=0.25\textwidth]{VQGAN-CLIP/generation/Baba_Yaga_s_house_fantasy_art.jpg}}\hfill
    \subfloat[pickled eggs, tempera on wood]{\includegraphics[width=0.25\textwidth]{VQGAN-CLIP/generation/eggs.jpg}}\\\
    \subfloat[effervescent hope]{\includegraphics[width=0.25\textwidth]{VQGAN-CLIP/generation/effervescent_hope.jpg}}\hfill
    \subfloat[the Tower of Babel by J.M.W. Turner]{\includegraphics[width=0.25\textwidth]{VQGAN-CLIP/generation/babel.jpg}}\hfill
    \subfloat[a futuristic city in synthwave style]{\includegraphics[width=0.25\textwidth]{VQGAN-CLIP/generation/synthwave.jpg}}
    \caption{Example \VC{} generations and their text prompts. This is the same figure as \cref{fig:example-gen} repeated for ease of comparison with \cref{fig:example-gen-min,fig:example-gen-clip,fig:example-gen-glide}.}
    \label{fig:example-gen-vc}
\end{figure}

\clearpage

\section{Comparison of Artistic Impressions}

We find that minDALL-E and both GLIDE (filtered) models have a much more tenuous understanding of famous artists than \VC{} does. To showcase this we invent six fictional but plausible painting titles by each artist in Figure 4 and prompt the models to generate, e.g., ``Willow Trees by van Gogh.'' While \VC{} is able to consistently produce something that is recognizably in the style of the artist and containing the subject, we find that the other models frequently fail to produce artwork that matches the prompt, either in subject or in style.

We note that the GLIDE (filtered) models will frequently produce photorealistic images, despite the fact that none of the artists produce photorealistic artwork. As mentioned in Section 4, our primary goal is to produce \textit{high quality art} rather than photorealistic images. Consequently, GLIDE's Hokusai generations - despite undoubtably being images of rice fields - rate lowly in our estimation as they have nothing in common with Hokusai's artwork. GLIDE's best results are in response to the van Gogh prompt in which the models clearly generate impressionistic images of trees, though the art resembles the work of Nicholas Verrall (in the case of CLIP-guided) and William Langson Lathrop or Claude Monet (CF-guided) far more so than van Gogh.

minDALL-E appears to be noticeably better at producing art, it frequently fails to agree with the prompt in both subject and in style. Perhaps most jarringly, in response to ``A Self-Portrait by Kahlo'' minDALL-E generates what is clearly a painting of a person, but the subject is a white man instead of a Latin woman.

\begin{figure*}
    \centering
    \subfloat{\includegraphics[width=0.33\textwidth]{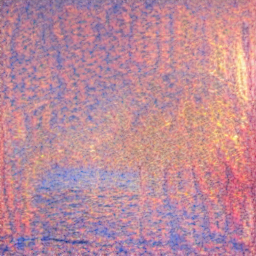}}\hfill
    \subfloat{\includegraphics[width=0.33\textwidth]{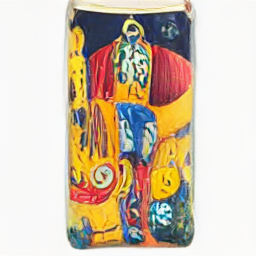}}\hfill
    \subfloat{\includegraphics[width=0.33\textwidth]{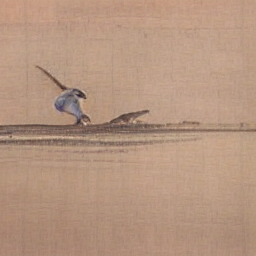}}\\
    \subfloat{\includegraphics[width=0.33\textwidth]{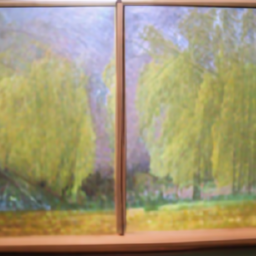}}\hfill
    \subfloat{\includegraphics[width=0.33\textwidth]{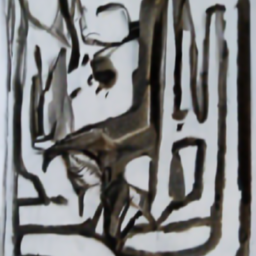}}\hfill
    \subfloat{\includegraphics[width=0.33\textwidth]{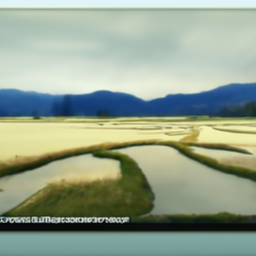}}\\
    \subfloat{\includegraphics[width=0.33\textwidth]{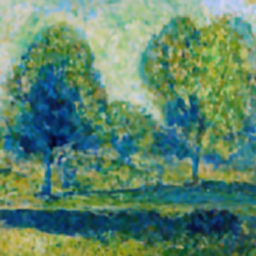}}\hfill
    \subfloat{\includegraphics[width=0.33\textwidth]{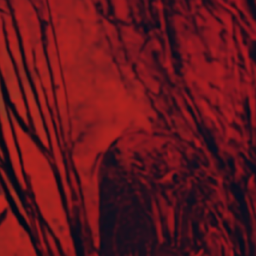}}\hfill
    \subfloat{\includegraphics[width=0.33\textwidth]{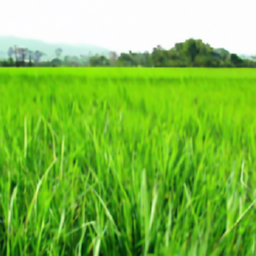}}\\
    \setcounter{subfigure}{0}
    \subfloat[Willow Trees by van Gogh]{\includegraphics[width=0.33\textwidth]{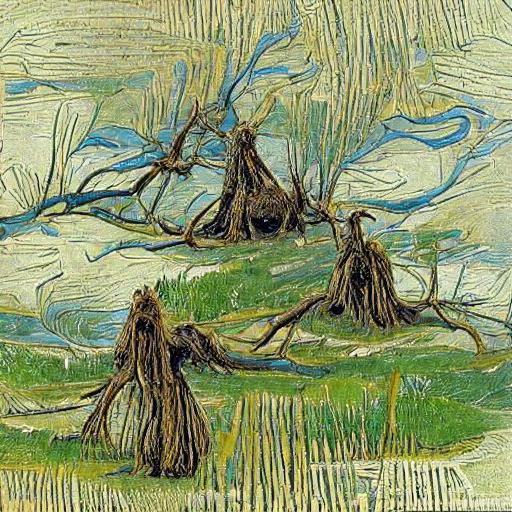}}\hfill
    \subfloat[The Woman by Picasso]{\includegraphics[width=0.33\textwidth]{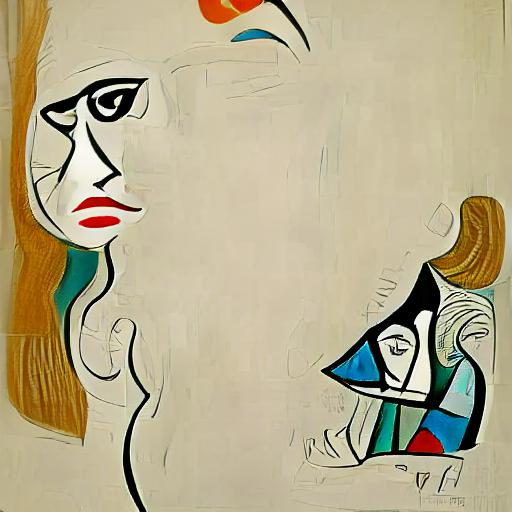}}\hfill
    \subfloat[Rice farming by Hokusai]{\includegraphics[width=0.33\textwidth]{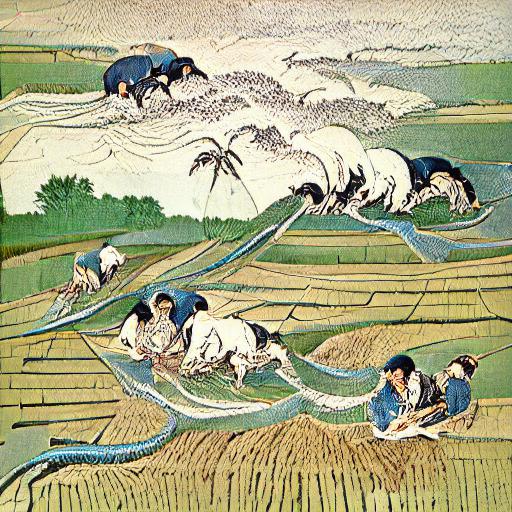}}\\
\caption{Attempts to generate novel works of art by famous artists. Top to bottom: minDALL-E, GLIDE (CLIP-guided), GLIDE (CF-guided), and our \VC{}. While VQGAN-CLIP's generation quality varies, it is the only model that seems to have grasped the desired task.}
\end{figure*}

\begin{figure*}
    \centering
    \subfloat{\includegraphics[width=0.33\textwidth]{VQGAN-CLIP/Famous Art/the_Tower_of_Babel_by_Turner_MIN}}\hfill
    \subfloat{\includegraphics[width=0.33\textwidth]{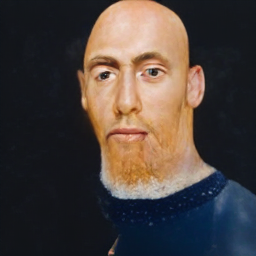}}\hfill
    \subfloat{\includegraphics[width=0.33\textwidth]{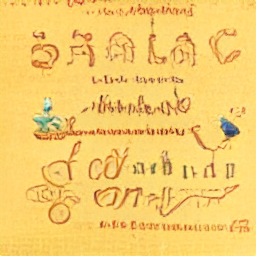}}\\
    \subfloat{\includegraphics[width=0.33\textwidth]{VQGAN-CLIP/Famous Art/the_Tower_of_Babel_by_Turner_CLIP}}\hfill
    \subfloat{\includegraphics[width=0.33\textwidth]{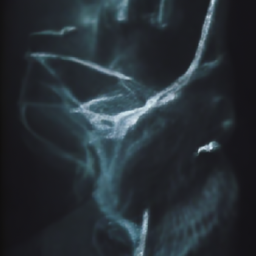}}\hfill
    \subfloat{\includegraphics[width=0.33\textwidth]{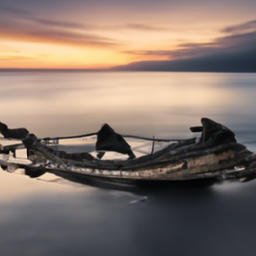}}\\
    \subfloat{\includegraphics[width=0.33\textwidth]{VQGAN-CLIP/Famous Art/The_Tower_of_Babel_by_Turner_GLIDE.png}}\hfill
    \subfloat{\includegraphics[width=0.33\textwidth]{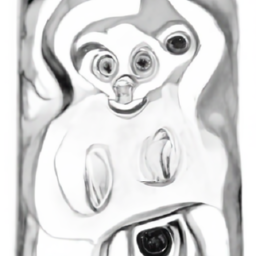}}\hfill
    \subfloat{\includegraphics[width=0.33\textwidth]{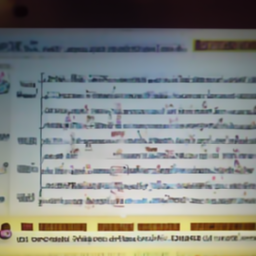}}\\
    \setcounter{subfigure}{0}
    \subfloat[the Tower of Babel by Turner]{\includegraphics[width=0.33\textwidth]{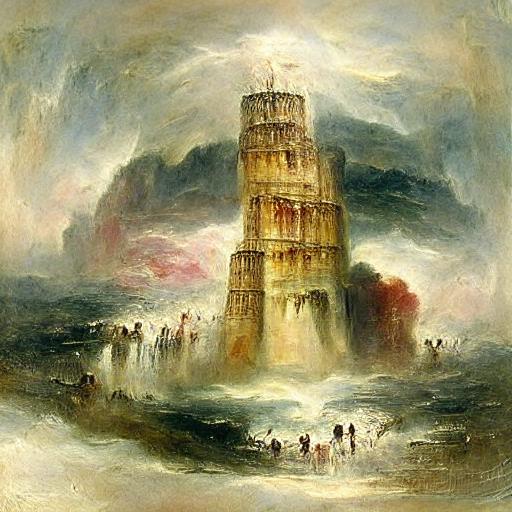}}\hfill
    \subfloat[A Self-Portrait by Kahlo]{\includegraphics[width=0.33\textwidth]{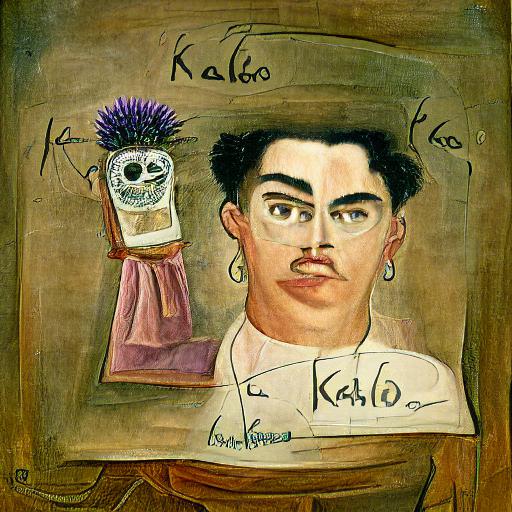}}\hfill
    \subfloat[In Search of Lost Time by Mehretu]{\includegraphics[width=0.33\textwidth]{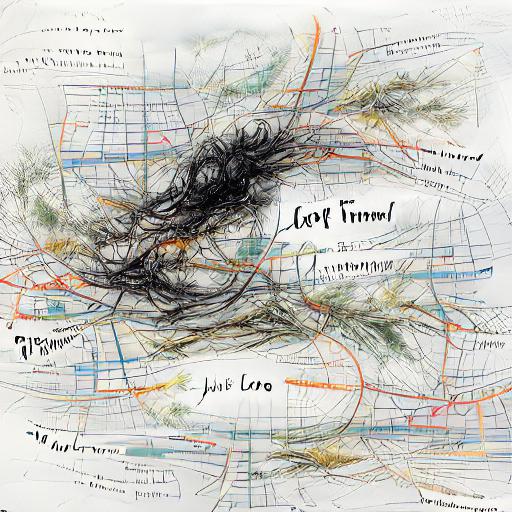}}\\
    \caption{Attempts to generate novel works of art by famous artists. Top to bottom: minDALL-E, GLIDE (CLIP-guided), GLIDE (CF-guided), and our \VC{}. While VQGAN-CLIP's generation quality varies, it is the only model that seems to have grasped the desired task.}
    \label{fig:famous-art-2}
\end{figure*}

As before, all images are cherrypicked best-of-five except for \VC{} which is uncherrypicked. We omit ablations over various wordings of the prompt as our exploratory testing indicated that it would effect the overall quality of the generation but not the extent to which they match the prompts.

\clearpage

\section{Additional Components}\label{app:more-components}

\subsection{Prompt Addition}\label{app:prompt-add}

\begin{figure}[!h]
    \centering
    \subfloat[A baseball game]{\includegraphics[width=0.2\textwidth]{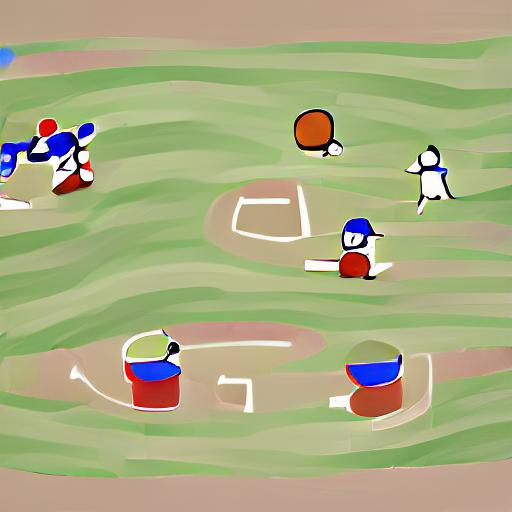}}\hfill
    \subfloat[A watercolor painting of a baseball game]{\includegraphics[width=0.2\textwidth]{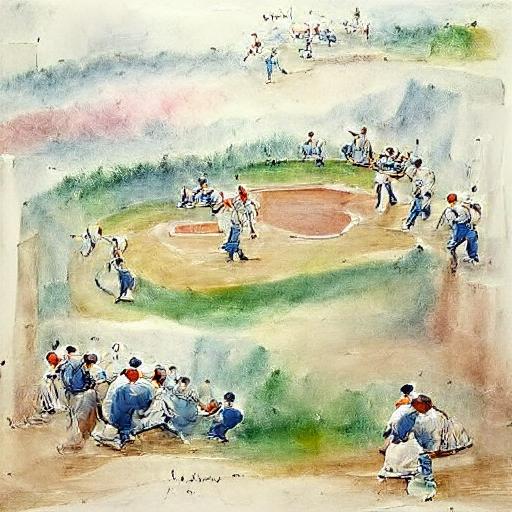}}\hfill
    \subfloat[A watercolor painting + A baseball game]{\includegraphics[width=0.2\textwidth]{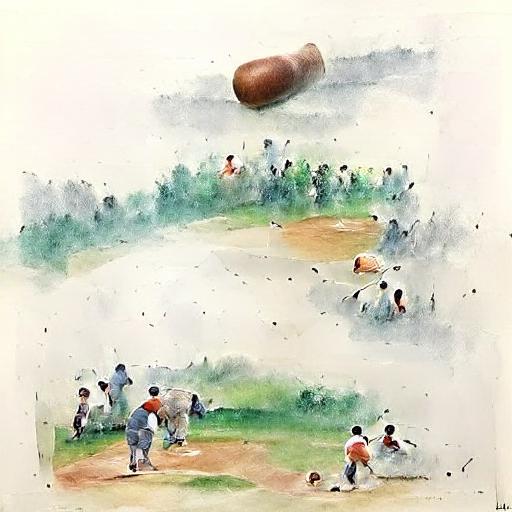}}\hfill
    \subfloat[A watercolor painting]{\includegraphics[width=0.2\textwidth]{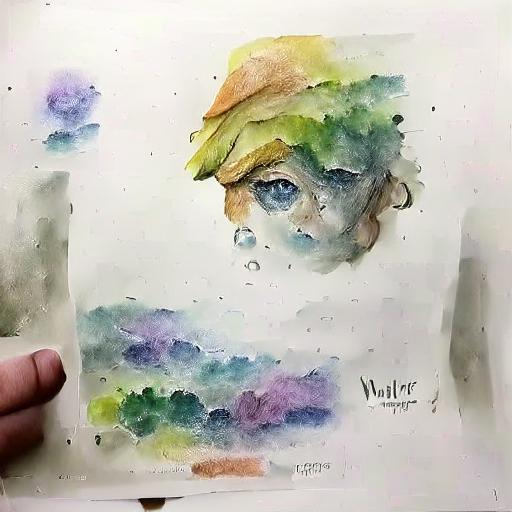}}\\
    \subfloat[A farm]{\includegraphics[width=0.2\textwidth]{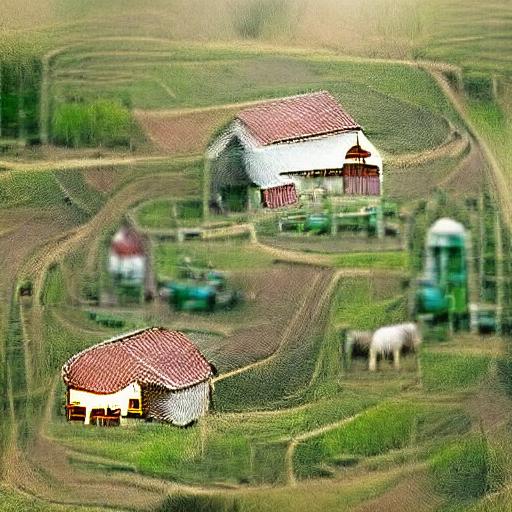}}\hfill
    \subfloat[A charcoal drawing of a farm]{\includegraphics[width=0.2\textwidth]{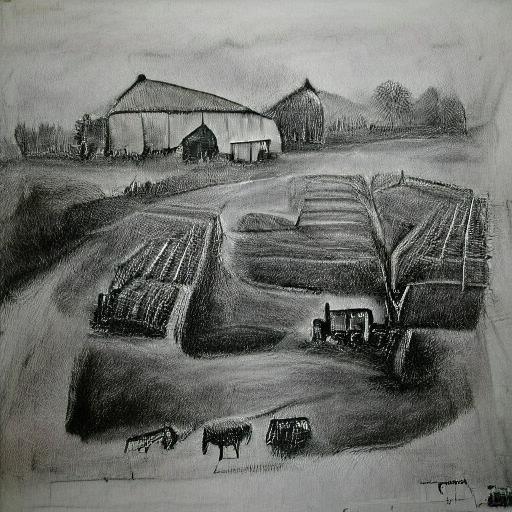}}\hfill
    \subfloat[A charcoal drawing + A farm]{\includegraphics[width=0.2\textwidth]{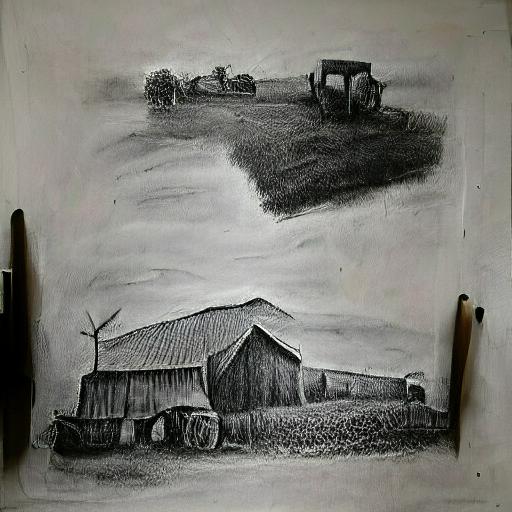}}\hfill
    \subfloat[A charcoal drawing]{\includegraphics[width=0.2\textwidth]{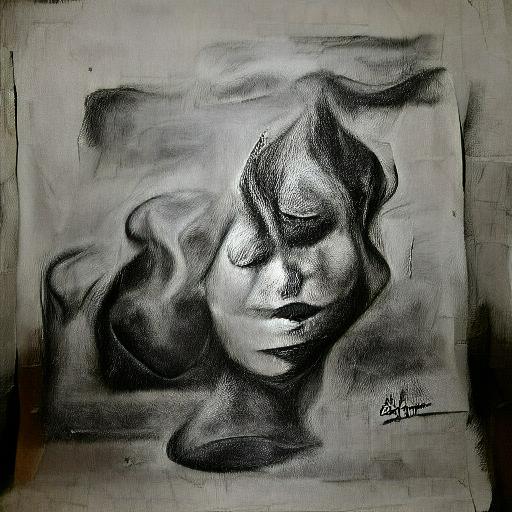}} 
    \caption{A comparison of natural semantic combination and addition in the latent space. Left to right: the original prompt, a natural language prompt, latent space addition of prompts, and the descriptive modifier.}
    \label{fig:latent-addition}
\end{figure}

One topic of immense interest for text-to-image models is \textit{compositionality}: the extent to which the model is able to take multiple discrete concepts and combine them. While a detailed analysis of compositionality in \VC{} is outside the scope of this paper, we have observed that \VC{} is able to intelligently combine multiple prompts. By providing two separate text inputs and averaging the loss, we can effectively ``add'' the two prompts together. In \cref{fig:latent-addition} we show that adding a content and a style prompt results in an image that is quite similar to one produced by combining the style and content into a single prompt with natural language. We view this as a rich and exciting area for future research.

\subsection{Masked Image Editing}\label{app:masking}

Given a source phrase (text) $s_p$ and a target phrase (text) $t_p$, we aim to construct an optimization rule that replaces all instances of $s_p$ within a source image $s_i$ with $t_p$, resulting in a target image (which is unknown at the beginning) $t_i$. 

To generate $t_i$, we first need to mask the part component of our source image that corresponds to $s_p$, for instance if $s_p$ is ``Dog,'' we need to mask the part of the image containing a dog. We can utilize CLIP in a zero shot setting to perform masking as follows:

\begin{enumerate}
    \item Crop $s_i$ into a number of smaller sub images, $\tilde{S} = \{S_i^k\}_{k=1}^N$
    \item $\forall i \in \tilde{S}$, compute $L(i) = f(i) \cdot g(s_p)$, where $f(\cdot)$ denotes our image encoder and $g(\cdot)$ denotes our text encoder.
    \item Record the value of L(i) at the center of the crop of $i$
    \item Normalize the resulting grey scale image, this is now our mask
\end{enumerate}

\begin{figure*}
    \centering
    \subfloat{\includegraphics[width=0.3\textwidth]{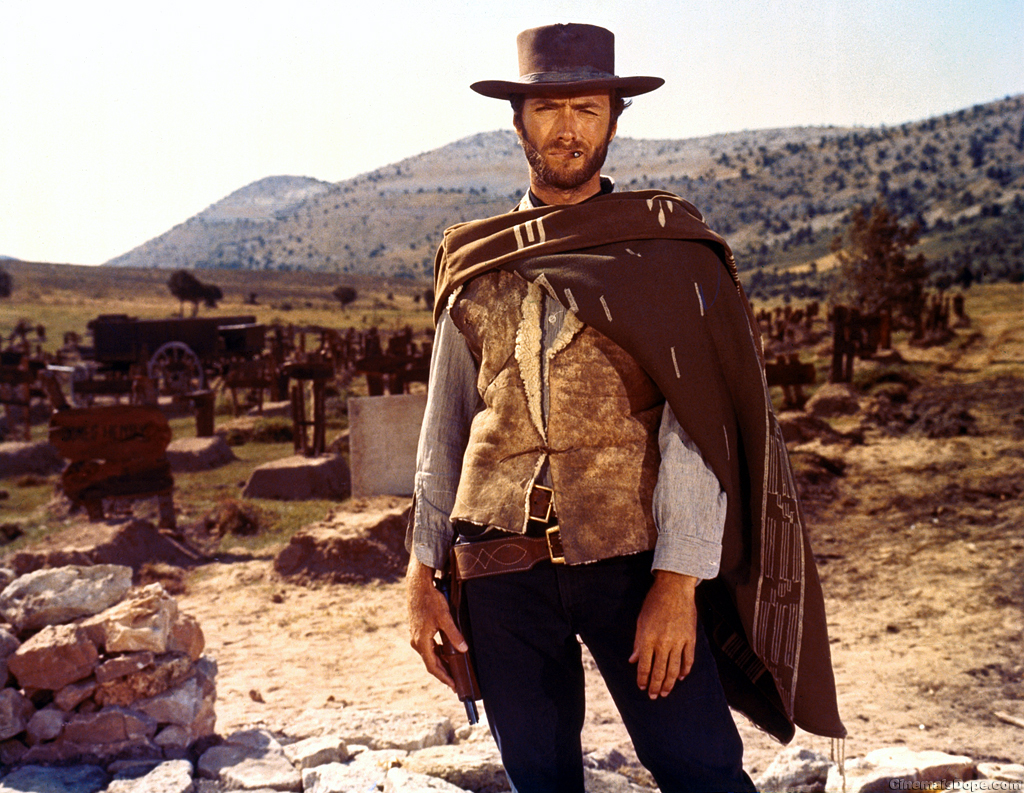}}\hfill
    \subfloat{\includegraphics[width=0.33\textwidth]{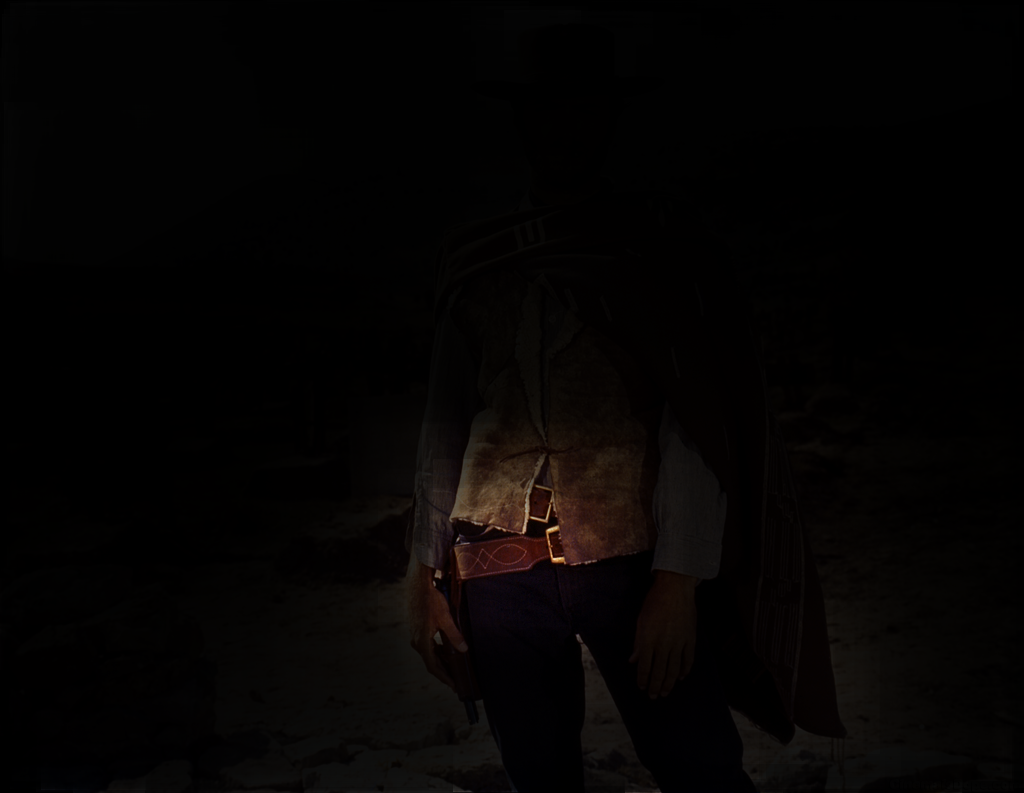}}\hfill
    \subfloat{\includegraphics[width=0.3\textwidth]{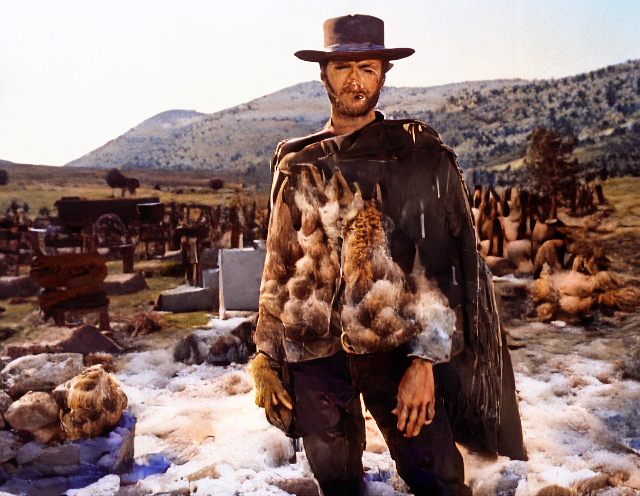}}\\
    \subfloat{\includegraphics[width=0.3\textwidth]{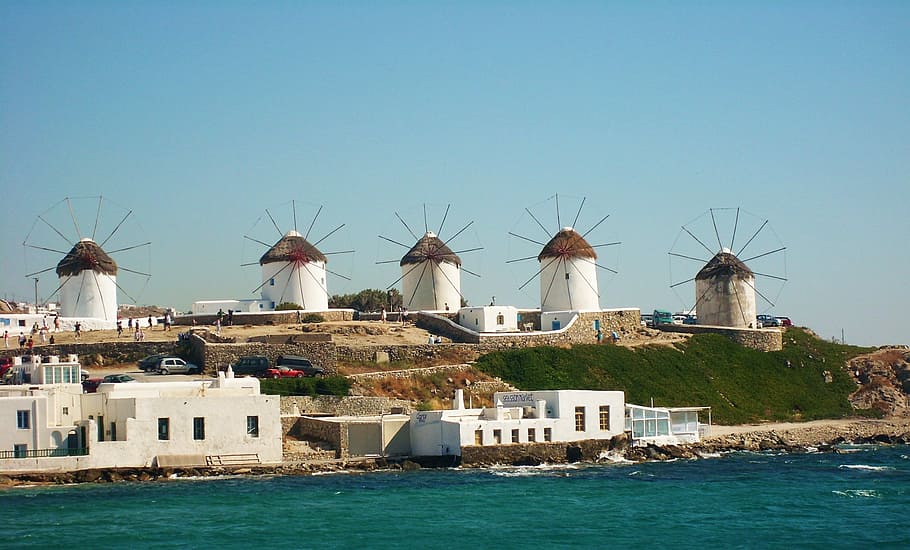}}\hfill
    \subfloat{\includegraphics[width=0.33\textwidth]{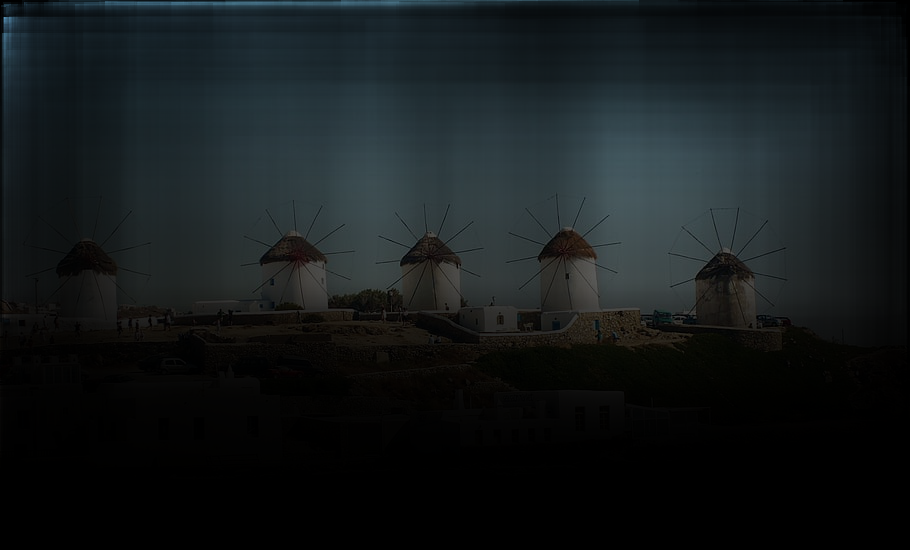}}\hfill
    \subfloat{\includegraphics[width=0.3\textwidth]{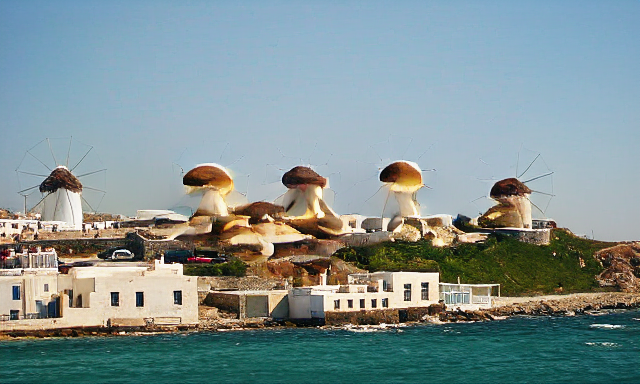}}\\
    \caption{Two examples of masked editing, showing Jacket $\to$ Fur and Windmills $\to$ Mushrooms.}
    \label{fig:editing}
\end{figure*}

We can then threshold this mask in order to determine which components of the mask actually contain our object of interest. In practice, we compute the threshold as two standard deviations below the average weight.

During generation, we crop and augment our current generated image and for every crop we compute the distance of the embedding of this crop to the embedding of the original image. This is used to preserve some notion of structure that was present in the original image. Similarly, we compute the distance of this embedded crop against the target phrase. 

By minimizing a weighted sum of these distances, we can perform in-image object replacement and editing. See Figure $\ref{fig:editing}$.
\end{document}